%% file: main.tex
\documentclass[10pt, a4paper, logo]{googledeepmind}

\usepackage{times}

\usepackage{hyperref}
\usepackage[leftcaption]{sidecap} 

\input{math_commands.tex}

\usepackage{hyperref}
\usepackage{url}
\usepackage{booktabs}
\usepackage{graphicx}
\usepackage{subcaption}
\usepackage{amssymb}
\usepackage[most]{tcolorbox}
\usepackage[table]{xcolor}
\tcbuselibrary{breakable} 
\usepackage{amsmath}
\usepackage{siunitx}
\usepackage{colortbl}

\usepackage{microtype}
\usepackage{color}
\usepackage{wrapfig}
\usepackage{natbib}
\usepackage{colortbl}
\usepackage{microtype}
\usepackage{graphicx}
\usepackage{booktabs} 
\usepackage{array}
\usepackage{textcomp}
\usepackage{stfloats}
\usepackage{float}
\usepackage{verbatim}
\usepackage[ruled, linesnumbered, lined]{algorithm2e}
\usepackage{multirow}
\usepackage{enumitem}

\usepackage{xcolor}

\definecolor{BestColor}{HTML}{C8E6C9}  
\definecolor{SecondBestColor}{HTML}{FFF9C4} 

\usepackage{tcolorbox}
\usepackage{amsmath,amsfonts}

\definecolor{ggg}{RGB}{26,179,0}
\definecolor{rrr}{RGB}{179,0,0}
\definecolor{oodc}{RGB}{31,73,121}
\definecolor{idc}{RGB}{68,142,68}

\definecolor{mygray}{gray}{0.9}

\def\Bias#1#2{\bm{b}}

\newtcolorbox{examplebox}[2][]{ 
    breakable, 
    enhanced, 
    colback=white, 
    colframe=cyan, 
    coltitle=white, 
    fonttitle=\bfseries, 
    title=#2, 
    overlay middle={\draw[cyan, line width=1pt](frame.south west)--(frame.south east);}, 
    overlay last={\draw[cyan, line width=1pt](frame.south west)--(frame.south east);}, 
    #1 
}

\usepackage[T1]{fontenc}
\usepackage{booktabs}      
\usepackage{graphicx}      
\usepackage[table]{xcolor} 
\usepackage{siunitx}       
\usepackage{etoolbox}      
\usepackage[normalem]{ulem}     

\usepackage{nicematrix}

\definecolor{impcolor}{HTML}{2E8B57} 

\newcommand{\improvementstyle}[1]{$^{\textcolor{impcolor}{\tiny #1}}$}

\newcommand{\scoreimp}[2]{%
  \textbf{#1}%
  \ifstrequal{#2}{+0.0}{}{%
    \ifstrequal{#2}{0.0}{}{%
      \makebox[0pt][l]{\improvementstyle{#2}}%
    }%
  }%
}

\title{Skill Self-Play: Proactive Curricula for Self-Evolving Agents}

\title{Skill Self-Play: Pushing the Frontier of LLM Capability with Co-Evolving Skills}

\author[1,2]{Siyuan Huang\textsuperscript{$\dag$}}
\author[1]{Pengyu Cheng\textsuperscript{$\S$}}
\author[1,3]{Haotian Liu\textsuperscript{$\dag$}}
\author[1,4]{Tao Chen\textsuperscript{$\dag$}}
\author[1,5]{Yihao Liu\textsuperscript{$\dag$}}
\author[1,6,7]{Jingwei Ni\textsuperscript{$\dag$}}
\author[1,8]{Shijie~Zhou\textsuperscript{$\dag$}}
\author[1]{Ziyi Yang}
\author[1]{Gangwei Jiang}
\author[1]{Mengyu Zhou}
\author[2]{Yu Cheng\textsuperscript{$\S$}}
\author[1]{Xiaoxi Jiang}
\author[1]{Guanjun Jiang}
\affil[1]{Qwen Large Model Application Team, Alibaba}
\affil[2]{The Chinese University of Hong Kong}
\affil[3]{Renmin~University~of~China}
\affil[4]{Sun~Yat-sen~University}
\affil[5]{Peking~University}
\affil[6]{ETH~Zürich}
\affil[7]{University~of~Zurich}
\affil[8]{University~at~Buffalo}

\input{Sections/0_Abstract}

\begin{document}
\maketitle

\input{Sections/1_Introduction}
\input{Sections/5_Related_Work}
\input{Sections/3_Methodology}
\input{Sections/4_Experiments}
\input{Sections/6_Conclusion}

\bibliography{conference}
\bibliographystyle{conference}

\appendix
\clearpage

\input{Sections/X_Appendix}

\end{document}

%% file: math_commands.tex
\usepackage{amsmath,amsfonts,bm}
\usepackage{natbib}
\usepackage{fancyhdr}

\def\eqref#1{equation~\ref{#1}}

\def\1{\bm{1}}

\def\vc{{\bm{c}}}

\def\vs{{\bm{s}}}

\def\vx{{\bm{x}}}
\def\vy{{\bm{y}}}

\DeclareMathAlphabet{\mathsfit}{\encodingdefault}{\sfdefault}{m}{sl}
\SetMathAlphabet{\mathsfit}{bold}{\encodingdefault}{\sfdefault}{bx}{n}

\def\gR{{\mathcal{R}}}
\def\gS{{\mathcal{S}}}



%% file: Sections/0_Abstract.tex
\begin{abstract}
LLM training is shifting from manual design and annotation to interaction-driven self-evolution. However, existing self-evolutionary methods face a fundamental dilemma between task diversity and verification reliability: environment-bound methods obtain precise feedback but confine learning to narrow domains, while open-ended self-generation broadens the task space but lacks reliable verification, allowing misleading rewards to pollute the training loop. We identify agent skills as a powerful middle ground to reconcile this tension: each skill ensures deep, verifiable execution in a specific scenario, while dynamic routing across skills maintains open-ended task variety.
Leveraging this insight, we introduce Skill Self-Play (Skill-SP), a co-evolutionary framework comprising a proposer, a solver, and a dynamic skill controller. Orchestrated via a reinforcement learning loop, these components co-evolve in a continuous self-play loop: the proposer generates challenging tasks conditioned on dynamically sampled skills; the solver explores candidate solutions to push its capability boundaries; and the skill controller collects execution feedback to update and expand the skill library. This interactive co-evolution effectively bridges the gap between structured verification and open-ended exploration. Empirical evaluations on tool-use and reasoning benchmarks demonstrate that Skill-SP, serving as a robust evolution engine, consistently pushes the performance ceiling of competent backbones while catalyzing striking turnarounds for initially misaligned models. Our code is available at \url{https://github.com/Qwen-Applications/skill-self-play}.
\end{abstract}

%% file: Sections/1_Introduction.tex
\begin{figure}[h]
    \centering
    \includegraphics[width=0.98\linewidth]{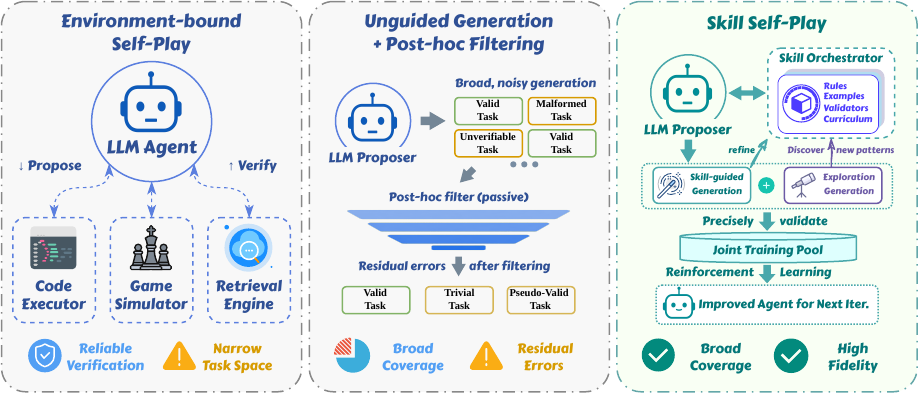} 
    
    \caption{{Motivation of \textbf{Skill Self-play}.}
    \textbf{Left:} \emph{Environment-bound methods} provide reliable verification but severely restrict the task space.
    \textbf{Middle:} \emph{Unguided generation} broadens coverage but depends on passive post-hoc filtering, allowing residual errors to degrade data fidelity.
    \textbf{Right:} \emph{Skill Self-Play} uses a proactive Skill Orchestrator with dual streams for skill-guided generation and open-ended exploration, enabling iterative skill evolution with both broad coverage and high fidelity.}

    \label{fig:intro}
\end{figure}

\section{Introduction}
\label{sec:introduction}

\begin{wrapfigure}{r}{0.39\textwidth}
    \vspace{-1.5em}
    \centering
    \includegraphics[width=\linewidth]{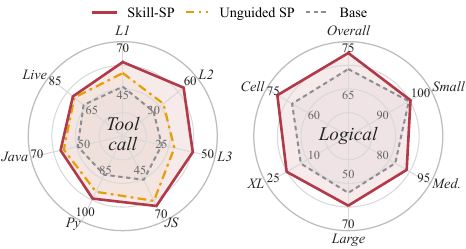}
    \vspace{-1.8em}
    \caption{\textbf{Performance footprint.} Skill-SP broadly expands Qwen3-4B-Ins capabilities across tool-calling and logical reasoning. Notably, Unguided SP is absent from the latter as it fails to synthesize valid puzzles.}
    \vspace{-1.0em}
    \label{fig:intro_radar}
\end{wrapfigure}

Self-evolution methods, which empower Large Language Models (LLMs) to autonomously generate tasks, find solutions, and evaluate outcomes, have shifted the post-training paradigm from heavy data-driven annotation to self-improvement without any supervision~\citep{tao2024survey,patel2024large,jiang2024long,chen2025multi,xu2025genius}. Along this new scaling dimension, self-play has emerged as a foundational technique, in which the learning LLM engages in gameplay as different role-playing agents and updates its policies via multi-agent reinforcement learning (MARL)~\citep{yang2025spell,yuan2025marshal,liu2025spiral,jiang2026one}. Recent works have demonstrated the surprising power of self-play across various scenarios, significantly boosting LLM capabilities in domains such as logical reasoning~\citep{cheng2024self,huang2025r}, deep search~\citep{lu2025search,liu2025spice}, visual question answering~\citep{wang2025vision}, and software engineering~\citep{wei2025toward,zhao2026absolute}.

However, the success of self-play in each specific domain depends heavily on well-crafted environments and rewards to provide precise validation~\citep{yang2026self,song2026envscaler,yan2026openskill,huang2026ace,wang2026code}. 
While external verification ensures accuracy, it confines agent training to narrow operational boundaries, leaving the model unable to generalize to broader, open-ended tasks~\citep{tang2025beyond,zhao2025learning}. 
To broaden the task space, alternative approaches employ unguided task generation paired with post-hoc filtering (e.g., format checks or majority voting)~\citep{wang2023self,xu2024wizardlm,nguyen2024better,xu2025magpie,an2025ultraif,yu2025cot}. 
Unfortunately, open-ended task generation severely compromises quality control~\citep{long2024llms,jiang2025self}. 
Without structural guidance, task generators frequently synthesize ill-posed tasks or overfit to simple templates~\citep{yu2025guided,li2026r}.
Over successive training iterations, accumulated errors and biases cause synthetic data collapse~\citep{shumailov2024ai,kazdan2025collapse}.
Ultimately, post-hoc filtering acts merely as a \emph{passive sieve}: it rejects glaring formatting errors but cannot actively guide the generator to produce reusable, logically valid, and diverse task patterns. 
Consequently, a fundamental challenge remains: \emph{How can self-play autonomously generate and verify diverse agent tasks without collapsing into either narrow domains or chaotic synthetic noise?}

To resolve this dilemma and bridge the gap between rigid environments and passive filtering~\citep{khattab2023dspy,dong2025self,prabhakar2026apigen,gao2026self}, we argue that verifiable self-play can be significantly enhanced by the concept of \emph{skills}. 
As introduced by~\citet{claudeIntroducingAgent}, a skill is a modular unit of procedural knowledge that bundles task-specific instructions and supporting resources into a reusable block. By organizing expertise into these self-contained units, skills allow agents to dynamically load specialized knowledge exactly when needed.
When formalized as a \emph{task-pattern interface}, this proactive abstraction serves as the ideal medium of evolution. 
While raw execution rollouts in self-play contain valuable insights, this evolutionary knowledge is typically scattered across isolated trajectories. 
Simply aggregating these experiences by stuffing historical trajectories into a single task-generator prompt leads to severe context bloat and dilutes structural control. 
In contrast, a skill interface distills this scattered experience into a compact, reusable, and co-evolvable unit. 
Rather than acting as a passive post-hoc filter, this interface actively orchestrates the entire self-play loop: injecting structural priors to guide the task generator \emph{before} synthesis, providing explicit executable mechanisms for rigorous verification \emph{after} response generation, and tracking historical difficulty to govern curriculum progression \emph{across} successive iterations.

To realize this paradigm, we introduce \emph{\textbf{Skill} \textbf{S}elf-\textbf{P}lay} (\textbf{Skill-SP}), a reinforcement learning framework driven by an evolving library of modular skill packages (Figure~\ref{fig:intro}). 
At its core, Skill-SP orchestrates self-play through three core components: a \emph{Proposer}, a \emph{Solver}, and a \emph{Controller}. 
More specifically, the Proposer synthesizes targeted challenges guided by routed skill packages; the Solver learns to execute and solve these tasks; and the Controller analyzes execution trajectories to continually manage the skill library. 
Within this loop, the skill library undergoes a continuous evolutionary cycle: the Controller automatically refines existing skill packages based on execution feedback, archives obsolete ones, and induces entirely new skills from open-ended exploration. 
This dynamic routing of modular skills ensures strict structural quality control without context bloat, while the discovery of new skill packages continuously expands the task frontier. 
By isolating governance in this evolving library, Skill-SP successfully resolves the fundamental tension between task diversity and verification reliability.
We empirically validate Skill-SP on two families of verifiable agent tasks: tool calling (API-Bank~\citep{li2023api} and BFCL~\citep{patil2025berkeley}) and logical reasoning (ZebraLogic~\citep{lin2025zebralogic}). Across various open-source LLM backbones, Skill-SP delivers substantial performance breakthroughs, yielding absolute gains of up to \textbf{+42.9} points on tool use and \textbf{+12.0} points on logical reasoning, consistently outperforming unguided self-play (as in Figure~\ref{fig:intro_radar}).
Finally, diagnostic analyses confirm that Skill-SP successfully anchors task generation at the model's evolving learning frontier by continuously distilling novel patterns into reusable skills. The proposed skill-driven self-play brings a fundamental paradigm shift to model self-improvement, unlocking a sustainable and truly open-ended frontier for LLM self-evolution.

%% file: Sections/5_Related_Work.tex
\section{Related Work}
\label{sec:related_work}

\paragraph{Self-play with external verifiers.}
Self-play has long improved agents in closed games~\citep{silver2017mastering,silver2018general}, inspiring recent language-model work on alignment, instruction following, and reasoning through self-generated data or verifiable rewards~\citep{cheng2024self,chen2024self,cheng2024spar,huang2025spotlight,kuba2025language,ren2026seif}. Recent self-evolving systems often obtain reliable feedback by grounding task generation in external environments. For example, Absolute Zero~\citep{zhao2026absolute} and verified-code pipelines~\citep{wilf2025propose} rely on code executors; SPIRAL and MARSHAL~\citep{liu2025spiral,yuan2025marshal} use game simulators; and Search Self-play~\citep{lu2025search} leverages retrieval augmentation. While these works demonstrate the value of reliable verification, their specialized environments inherently bottleneck the task distribution, confining the agent's learning to narrow, predefined operational domains.

\paragraph{Synthetic task generation and filtering.}
To broaden the task space, general self-improvement pipelines commonly combine synthetic task generation with automated verification~\citep{yang2026self}. Representative mechanisms include unit tests for coding tasks~\citep{jimenez2024swe,gehring2024rlef,jiang2025coderl+}, success signals in interactive web environments~\citep{liu2024agentbench,zhou2024webarena,yao2024tau,song2026envscaler}, schema checks for tool use~\citep{li2023api,patil2025berkeley}, and deterministic constraint checkers for reasoning~\citep{zhou2023instruction,jiang2024followbench,qin2024infobench,wen2024benchmarking,pyatkin2025generalizing}. Although these mechanisms make synthetic data auditable, they primarily operate as passive, \emph{post-hoc} filters. They dictate which tasks survive but provide no proactive structural guidance to the proposer, causing unguided generation to frequently collapse into ill-posed tasks or redundant templates.

\paragraph{Skill interfaces for agents.}
Recent work studies agent skills as reusable procedural interfaces, with an emphasis on retrieval, compression, and progressive disclosure~\citep{ling2026agent,jiang2026sok,cho2026skillret,li2026skillsinjector,huang2026persistent,gao2026skillreducer,chen2026skilljuror,chen2026skill,he2026gems}. Another line of research automatically constructs or evolves skills from interaction traces and execution feedback~\citep{mi2026procmem,liu2026harnessing,wang2026skillx,shen2026skillfoundry,yang2026skillmaster,liu2026skillrevise,gautam2026skillaxe,wang2026skillgrad,liu2026skillsvote,ni2026trace2skill}, although preserving the fidelity of automatically generated skills remains challenging~\citep{li2026skillsbench,zhong2026skilllearnbench,zhou2026skillgenbench}. Most prior work uses skills to support inference-time execution, while a few recent studies explore their role during training~\citep{lu2026skill0}. In contrast, Skill-SP uses an evolving skill library as a training-time interface for task synthesis, enabling structural guidance and verification to co-evolve with the self-play curriculum. The following section formalizes this framework.

%% file: Sections/3_Methodology.tex
\section{Methodology}
\label{sec:methodology}

\begin{figure}[t]
    \centering
    \includegraphics[width=0.98\linewidth]{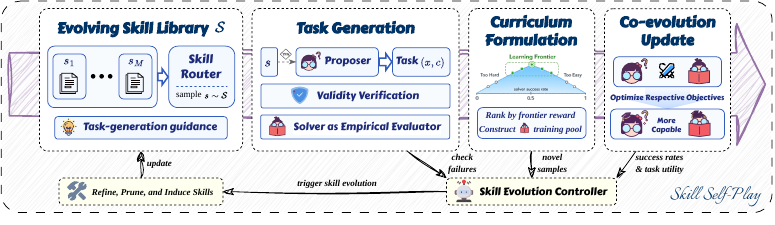} 
    \vspace{-0.5em}
    \caption{\textbf{Overview of Skill Self-Play.} An evolving skill library routes task-generation guidance to the proposer, which generates candidate tasks that undergo validity verification. Valid candidates are ranked by frontier reward to construct the solver curriculum. The resulting signals drive co-evolutionary proposer and solver updates, while validation failures, novel samples, and task-level statistics trigger skill refinement, pruning, and induction, distilling training feedback into reusable skills for subsequent task generation.}
    \label{fig:method}
\end{figure}

We propose \emph{\textbf{Skill} \textbf{S}elf-\textbf{P}lay} (\textbf{Skill-SP}), a training-time framework that transforms generic self-play into a proactive curriculum-construction process (Figure~\ref{fig:method}). At each iteration, Skill-SP orchestrates the joint optimization of a proposer policy $\pi_\text{propose}$ and a solver policy $\pi_\text{solve}$, alongside a skill controller managing an evolving library $\gS$. Guided by routed skills, the proposer is updated to synthesize valid tasks at the solver's learning frontier, while the solver is optimized on the resulting verified curriculum. Concurrently, the controller distills execution feedback to update the library, driving a continuous co-evolutionary loop that yields progressively more informative challenges.

\subsection{Skill Self-Play Objective}
\label{sec:osp_objective}

We formalize a verifiable agent task as a tuple $(\vx,\vc)$, where $\vx$ is the standard prompt visible to the solver, and $\vc$ is a hidden, machine-readable verification contract (e.g., unit tests, reference answers) evaluated exclusively by the environment. Given a solver's response $\vy$, the environment returns a verification reward $\mathcal{R}_\text{solve}(\vx,\vy,\vc) \in [0,1]$.

A standard self-play loop trains the proposer to generate challenging tasks and the solver to resolve them. For a generated task instance $(\vx,\vc)$, the solver's expected success rate is defined as:
\begin{equation}
v_\text{solve}(\vx,\vc;\pi_\text{solve})
=
\mathbb{E}_{\vy \sim \pi_\text{solve}(\cdot \mid \vx)}
\left[
    \mathcal{R}_\text{solve}(\vx,\vy,\vc)
\right].
\label{eq:osp_empirical_success}
\end{equation}
To drive continuous improvement, a natural objective for the proposer is to target the solver's learning frontier via a medium-difficulty score, formulated as $1 - 2|v_\text{solve} - 0.5|$. However, optimizing the proposer solely toward this metric frequently invites \emph{reward hacking}, where the proposer synthesizes ill-posed or unsolvable contracts to fabricate artificial difficulty.
To prevent this, Skill-SP formalizes task generation via a \emph{gated curriculum reward}. We introduce a binary task quality filter that retains only candidates with structurally sound contracts, thereby explicitly gating the proposer's reward:
\begin{equation}
\gR_\text{propose}(\vx,\vc;\pi_\text{solve})
=
\mathbb{1}_{\{(\vx,\vc) \text{ is valid}\}}
\cdot
\big(
    1 - 2
    \Big|
        v_\text{solve}(\vx,\vc;\pi_\text{solve})
        - \frac{1}{2}
    \Big|
\big),
\label{eq:frontier_score}
\end{equation}
where the indicator $\mathbb{1}_{\{(\vx,\vc) \text{ is valid}\}}$ represents this binary filter, which is further detailed in Eq.~\ref{eq:valid-task-indicator} of Section~\ref{sec:proposer_optimization}.

To proactively guide task generation, Skill-SP introduces an evolving skill library $\gS$. Task synthesis is conditioned on a sampled skill $\vs \sim \gS$, which equips the proposer with reusable structural priors and programmatic validators. Consequently, Skill-SP can be formulated as a bi-level optimization problem. The outer objective jointly optimizes the skill library $\gS$ and the proposer $\pi_\text{propose}$ to synthesize valid, frontier-targeted tasks, subject to an inner objective where the solver maximizes its execution success on these tasks:
\begin{equation}
\begin{aligned}
    \max_{\pi_\text{propose}, \gS} \quad
    & \mathbb{E}_{\vs \sim \gS,\;(\vx,\vc)\sim \pi_\text{propose}(\cdot\mid\vs)}
    \left[\gR_\text{propose}(\vx,\vc;\pi^*_\text{solve})\right] \\
    \text{s.t.} \quad
    &
    \pi^*_\text{solve}
    =
    \arg\max_{\pi}
    \mathbb{E}_{
        \vs\sim \gS,\; (\vx,\vc)\sim\pi_\text{propose}(\cdot\mid\vs),\;
        \vy\sim\pi(\cdot\mid\vx)
    }
    \left[
        \mathcal{R}_\text{solve}(\vx,\vy,\vc)
    \right].
\end{aligned}
\label{eq:bilevel_optimization}
\end{equation}
To orchestrate this high-fidelity curriculum loop, the subsequent sections detail the policy optimization of the proposer and the solver, alongside the dynamic evolution of the skill library $\gS$.

\subsection{Proposer Optimization via Skill-Conditioned Generation}
\label{sec:proposer_optimization}

To operationalize the outer objective (Eq.~\ref{eq:bilevel_optimization}), the proposer must continuously synthesize valid, frontier-targeted tasks. We structure this process into three stages: skill-conditioned generation to inject structural priors, rigorous verification to ensure task validity, and policy optimization to drive curriculum evolution.

\textbf{Skill-Conditioned Generation.}
A skill $\vs \in \gS$ serves as a structural interface defined by the tuple $\vs = \langle m, r, h, e, \nu, \sigma \rangle$. Here, $m$ denotes routing metadata; $r$, $h$, and $e$ provide procedural rules, generation hints, and few-shot examples to contextually condition the proposer; $\nu$ specifies executable validators; and $\sigma$ tracks historical usage statistics. During generation, we dynamically sample a skill $\vs \sim \gS$ based on $\sigma$, balancing the exploitation of high-yield skills with an exploration bonus for under-tested ones (detailed in Appendix~\ref{app:skill_routing}). The proposer then synthesizes candidates via a \emph{skill stream} $(\vx,\vc) \sim \pi_\text{propose}(\cdot \mid \vs)$, which injects strong structural priors. To continuously chart novel task spaces and prevent mode collapse, this is complemented by an open-ended \emph{exploration stream} $(\vx,\vc) \sim \pi_\text{propose}(\cdot \mid \varnothing)$ without skill constraints.

\textbf{Validity Verification.}
The binary task quality filter $\mathbb{1}_{\{(\vx,\vc) \text{ is valid}\}}$ gating the proposer reward (Eq.~\ref{eq:frontier_score}) enforces rigorous structural and logical checks. For tasks generated via the skill stream, this valid event is defined as the intersection of three conditions:
\begin{equation}\label{eq:valid-task-indicator}
    \big\{(\vx,\vc) \text{ is valid}\big\} = \big\{\text{schema compliant}\big\} \cap \big\{\text{contract valid}\big\} \cap \big\{\text{probe consistent}\big\}.
\end{equation}
Here, the first condition ensures global schema compliance; the second applies validators defined in $\vs$ to filter ill-posed contracts; and the third verifies \emph{probe consistency}. Specifically, this consistency requires that $K$ rollouts drawn from the current solver $\pi_\text{solve}(\cdot \mid \vx)$ yield a unique majority answer matching the proposer-defined reference in $\vc$. For the exploration stream, the valid event omits this skill-specific validation.

\textbf{Policy Update.}
At each training iteration $t$, we fix the current solver $\pi_{\text{solve}}^{(t)}$ to serve as an empirical evaluator. For each generated candidate, we compute the gated reward $\gR_\text{propose}$ by combining its validity mask $\mathbb{1}_{\{(\vx,\vc) \text{ is valid}\}}$ with the empirical success rate $v_\text{solve}(\vx,\vc;\pi_{\text{solve}}^{(t)})$, efficiently estimated by reusing the $K$ verification rollouts. The proposer policy $\pi_{\text{propose}}^{(t)}$ is then updated via Group Relative Policy Optimization (GRPO)~\citep{shao2024deepseekmath} to maximize the expected reward:
\begin{equation}
\mathcal{J}_{\text{propose}}(\pi)
=
\mathbb{E}_{\vs \sim \gS, \; (\vx,\vc) \sim \pi(\cdot \mid \vs)}
\left[
\gR_{\text{propose}}
(\vx,\vc;\pi_{\text{solve}}^{(t)})
\right].
\label{eq:proposer_rl_objective}
\end{equation}
Optimizing this objective yields the next-iteration proposer $\pi_{\text{propose}}^{(t+1)}$, ensuring that as the solver evolves, the proposer dynamically tracks its shifting learning frontier to continuously synthesize challenging, reliable tasks.

\subsection{Solver Optimization via Dynamic Curriculum Construction}
\label{sec:solver_optimization}

To operationalize the inner objective (Eq.~\ref{eq:bilevel_optimization}), the solver is optimized on a dynamic curriculum that systematically advances its capabilities. We construct this high-quality training pool by selecting the most challenging, frontier-targeted tasks from both generation streams.

\textbf{Curriculum Construction.}
Let $\mathcal{T}_{\text{skill}}^{(t)}$ and $\mathcal{T}_{\text{explore}}^{(t)}$ denote the sets of strictly valid candidates $(\vx,\vc)$ synthesized by the skill and exploration streams, respectively. To form the final training pool $\mathcal{D}^{(t)}$ of size $M$, we rank these candidates by their proposer reward $\gR_{\text{propose}}(\vx,\vc;\pi_{\text{solve}}^{(t)})$, effectively pinpointing the solver's learning frontier, and select the top-scoring tasks according to a blending ratio $\alpha \in (0,1)$:
\begin{equation}
    \mathcal{D}^{(t)}
    =
    \mathrm{Top}_{\alpha M} \! \left(\mathcal{T}_{\text{skill}}^{(t)}; \gR_\text{propose}\right)
    \; \cup \;
    \mathrm{Top}_{(1-\alpha) M} \! \left(\mathcal{T}_{\text{explore}}^{(t)}; \gR_\text{propose}\right).
    \label{eq:mixed_pool}
\end{equation}

\textbf{Policy Update.}
Using the constructed curriculum $\mathcal{D}^{(t)}$, the solver policy $\pi_{\text{solve}}^{(t)}$ is updated via GRPO to maximize the environment verification reward:
\begin{equation}
\mathcal{J}_{\text{solve}}(\pi)
=
\mathbb{E}_{
(\vx,\vc)\sim\mathcal{D}^{(t)},\;
\vy\sim\pi(\cdot\mid\vx)
}
\left[
\mathcal{R}_{\text{solve}}(\vx,\vy,\vc)
\right].
\label{eq:solver_rl_objective}
\end{equation}
Optimizing this objective yields the next-iteration solver $\pi_{\text{solve}}^{(t+1)}$.

\subsection{Continuous Evolution of the Skill Library}
\label{sec:skill_evolution}

To prevent curriculum stagnation, the skill library $\gS^{(t)}$ dynamically co-evolves with the proposer and solver policies at each iteration $t$. This evolution comprises three operations: refining existing skills, pruning obsolete ones, and inducing novel structural priors from open-ended exploration.

\textbf{Skill Refinement.}
Using generation feedback, Skill-SP updates the tracking statistics $\sigma$ of each sampled skill to modulate future routing (Appendix~\ref{app:skill_routing}). Concurrently, it analyzes execution trajectories from invalid generation attempts to diagnose systematic failures and automatically refine the core content of the skills, maintaining the library as a robust repository of structural priors.

\textbf{Skill Pruning.}
To maximize sample efficiency, the active library undergoes periodic pruning. By monitoring $\sigma$, Skill-SP identifies a pruning set $\mathcal{S}_{\text{prune}}^{(t)}$ of saturated skills that consistently yield trivial tasks, indicated by an expected frontier reward falling below a threshold $\gamma_{\text{prune}}$:
\begin{equation}
    \mathcal{S}_{\text{prune}}^{(t)} = \left\{ \vs \in \gS^{(t)} \;\middle|\; \mathbb{E}_{(\vx,\vc)\sim\pi_{\text{propose}}^{(t)}(\cdot\mid\vs)} \big[ \gR_{\text{propose}}(\vx,\vc;\pi_{\text{solve}}^{(t)}) \big] < \gamma_{\text{prune}} \right\}.
\end{equation}
Archiving $\mathcal{S}_{\text{prune}}^{(t)}$ conserves the generation budget and keeps the proposer focused on the learning frontier.

\textbf{Skill Induction and Update.}
To systematically expand the curriculum, Skill-SP extracts an induction set $\mathcal{T}^{(t)}_{\text{induce}}$ of novel candidates from the exploration stream, conditioned on a minimum frontier reward threshold $\gamma_{\text{induce}}$:
\begin{equation}
    \mathcal{T}^{(t)}_{\text{induce}} = \left\{ (\vx,\vc) \in \mathcal{T}_{\text{explore}}^{(t)} \;\middle|\; \gR_{\text{propose}}(\vx,\vc;\pi_{\text{solve}}^{(t)}) \ge \gamma_{\text{induce}} \right\}.
\end{equation}
Skill-SP then leverages the skill controller $\pi_{\text{control}}$, instantiated by the initial base policy, to abstract the underlying task patterns within $\mathcal{T}^{(t)}_{\text{induce}}$ into candidate packages. These are systematically filtered for structural integrity and semantic novelty against the current library to yield the newly induced skills $\gS_{\text{new}}^{(t)}$:
\begin{equation}
    \gS_{\text{new}}^{(t)} = \big\{ \vs \sim \pi_{\text{control}}( \cdot | \mathcal{T}^{(t)}_{\text{induce}}) \;\big|\; \operatorname{Integrity}(\vs) \wedge \operatorname{Novelty}(\vs, \gS^{(t)}) \big\},
\end{equation}
where $\operatorname{Integrity}(\vs)$ checks package completeness, and $\operatorname{Novelty}(\vs,\gS^{(t)})$ rejects candidates duplicating existing skills by thresholding lexical similarity.
The active library is subsequently updated via $\gS^{(t+1)} = (\gS^{(t)} \setminus \mathcal{S}_{\text{prune}}^{(t)}) \cup \gS_{\text{new}}^{(t)}$. This mechanism distills successful open-ended exploration into reusable structural priors, driving an ever-expanding curriculum. Algorithm~\ref{alg:skill_sp_overview} summarizes the resulting training loop.

\begin{algorithm}[t]
    \caption{{Skill Self-Play} (Skill-SP)}
    \label{alg:skill_sp_overview}
    \KwIn{Initial proposer $\pi_\text{propose}^{(0)}$, solver
    $\pi_\text{solve}^{(0)}$, skill library $\gS^{(0)}$, and number of
    self-play iterations $T$.}
    \KwOut{Trained solver $\pi_\text{solve}^{(T)}$.}

    \ForEach{self-play iteration $t$}{
        Sample $\vs\sim\gS^{(t)}$ using skill statistics; generate candidates with
        $\pi_\text{propose}^{(t)}(\cdot\mid\vs)$ and
        $\pi_\text{propose}^{(t)}(\cdot\mid\varnothing)$\;

        Verify candidates and compute $\gR_\text{propose}$; retain valid candidates for the solver curriculum\;

        Rank valid candidates by $\gR_\text{propose}$ to build the mixed
        frontier curriculum $\mathcal{D}^{(t)}$\;

        Update $\pi_\text{propose}^{(t)}\!\rightarrow\!
        \pi_\text{propose}^{(t+1)}$ using $\mathcal{J}_\text{propose}$, and
        $\pi_\text{solve}^{(t)}\!\rightarrow\!\pi_\text{solve}^{(t+1)}$
        on $\mathcal{D}^{(t)}$ using $\mathcal{J}_\text{solve}$\;

        Evolve the skill library by updating skill statistics, refining existing skills from failure traces, inducing new skills $\gS_{\text{new}}^{(t)}$ from
        induction candidates $\mathcal{T}_{\text{induce}}^{(t)}$, and pruning obsolete skills
        $\mathcal{S}_{\text{prune}}^{(t)}$:
        $\gS^{(t+1)}\gets
        (\gS^{(t)}\setminus\mathcal{S}_{\text{prune}}^{(t)})\cup
        \gS_{\text{new}}^{(t)}$\;
    }
    \Return{$\pi_\text{solve}^{(T)}$}\;
\end{algorithm}

%% file: Sections/4_Experiments.tex
\section{Experiments}
\label{sec:experiments}

\subsection{Experimental Setup}
\label{sec:experimental_setup}

\paragraph{Task Families.} 
We instantiate the framework on two domains. The first is \emph{tool-call prediction}, evaluating the solver's schema adherence and tool selection. We benchmark on API-Bank Levels 1--3~\citep{li2023api} and four BFCL categories~\citep{patil2025berkeley}: \textit{bfcl\_simple\_javascript}, \textit{bfcl\_simple\_python}, \textit{bfcl\_simple\_java}, and \textit{bfcl\_live\_simple}. The second is \emph{logical reasoning}, evaluating the solver's ability to satisfy complex constraints and deduce unique solutions. We benchmark on ZebraLogic~\citep{lin2025zebralogic}, which formalizes tasks as Zebra-style grid puzzles spanning four complexity scales determined by their search space size.

\paragraph{Backbones.}
We evaluate five backbones spanning 3B to 14B parameters: Qwen3-4B-Instruct, Qwen3-8B~\citep{yang2025qwen3}, Ministral-3-8B-Instruct, Ministral-3-14B-Instruct~\citep{liu2026ministral}, and Granite-4.1-3B~\citep{granite2026}. For each backbone, the identical checkpoint initializes both the proposer and the solver. Unless otherwise stated, skill refinement and induction rely solely on the backbone itself, without any external stronger teacher.

\paragraph{Baselines \& Evaluation Strategy.}
To isolate the end-to-end impact of our framework, the main results (Section~\ref{sec:main_results}) evaluate the final solvers trained via Skill-SP against both the initial \emph{Base} checkpoints and an \emph{Unguided SP} baseline, which performs standard self-play relying solely on passive post-hoc filtering without proactive skill guidance. Crucially, because unguided generation structurally collapses on complex tasks like synthesizing valid logical puzzles, Unguided SP fails to bootstrap a viable training loop in the reasoning domain and is therefore evaluated exclusively on tool calling. Further mechanistic diagnostics and comparisons against other controlled variants are detailed in the ablation study (Section~\ref{sec:ablations}).

\paragraph{Training Protocol \& Implementation Details.}
All experiments run for five iterations, optimizing policies via GRPO (4 proposer and 5 solver rollouts). The per-iteration training pool contains $M=8{,}000$ tasks ($\alpha=0.5$) via $K=10$ probes for tool calling, and $M=1{,}920$ tasks via a deterministic checker for reasoning. The proposer and solver undergo 5 and 15 update steps respectively for tool calling (3 and 10 for reasoning). Solver rewards target exact correctness and format compliance; the proposer receives a $-1$ penalty for structural parsing failures. Final benchmark results are reported as avg@8. The full hyperparameters, initial library $\gS^{(0)}$, and formal algorithm are detailed in Appendices~\ref{app:training_details}, \ref{app:initial_skills}, and \ref{app:osp_pseudocode}, respectively.

\input{tables/tool_call_main}

\subsection{Main Results}
\label{sec:main_results}

\paragraph{Tool-call prediction.}
Table~\ref{tab:tool_call_main} reports the end-to-end performance on tool calling. Skill-SP universally improves all five backbones across diverse tool-calling scenarios. For competent base models like Qwen and Granite, the framework yields steady average enhancements of 2.8 to 6.5 absolute points. Crucially, Unguided SP not only yields consistently smaller overall gains, but also behaves inconsistently, occasionally causing capability degradation on specific subtasks. In contrast, Skill-SP maintains strictly positive and superior gains across the board. More strikingly, Skill-SP rescues initially misaligned models from severe zero-shot schema adherence failures. On Ministral-3-8B, for example, it drives a massive 42.9-point absolute gain where Unguided SP remains completely stagnant. This stagnation occurs because the base model completely fails to synthesize valid tasks independently, starving the unguided loop of meaningful learning signals. These results confirm that proactive skill orchestration translates reliably into generalized tool-use capabilities.

\paragraph{Logical reasoning.}
Table~\ref{tab:logical_main} confirms that Skill-SP successfully transfers to rigorous constraint-satisfaction reasoning, universally improving overall accuracy across all five backbones. While competent models like Qwen3-4B-Instruct show steady enhancements across all task scales, the framework drives striking turnarounds for logically misaligned models such as Ministral-3-14B, delivering remarkable grid-level gains of up to 12.0 points overall and exceeding 35 points on smaller puzzles. Naturally, progress on the extreme Large and X-Large scales remains limited for initially weak models, as pure self-play requires a minimal capability threshold to bootstrap valid learning signals. This confirms that guaranteeing the structural accuracy and broad diversity of synthesized puzzles robustly translates into advanced logical deduction.

\input{tables/logical_main}

\subsection{Ablations}
\label{sec:ablations}

To isolate the contribution of each design choice, we evaluate controlled variants on Qwen3-4B-Instruct shown in Figure~\ref{fig:toolcall_data_efficiency} and Tables~\ref{tab:ablations} and \ref{tab:frozen_component_ablations_best}. The results systematically validate four core design dimensions:

\begin{wrapfigure}{r}{0.67\columnwidth}
    \vspace{-1.5em}
    \centering
    \begin{minipage}[t]{0.49\linewidth}
        \centering
        \includegraphics[width=\linewidth]{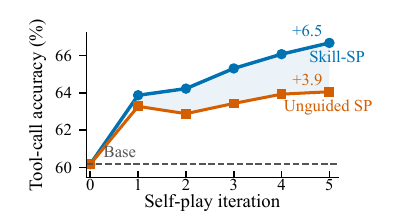}\\[-0.5em]
        {(a) Learning trajectory.}
    \end{minipage}
    \hfill
    \begin{minipage}[t]{0.49\linewidth}
        \centering
        \includegraphics[width=\linewidth]{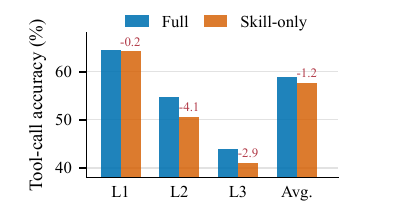}\\[-0.5em]
        {(b) Skill-only ablation.}
    \end{minipage}
    \vspace{-0.6em}
    \caption{\textbf{Tool-call ablations.} Skill-SP improves more steadily than Unguided SP, while skill-only data underperforms the mixed pool.}
    \label{fig:toolcall_data_efficiency}
    \vspace{-1.5em}
\end{wrapfigure}

\textbf{Naive self-play.} The \emph{Unguided SP} variant removes skill orchestration entirely. As depicted in Figure~\ref{fig:toolcall_data_efficiency}a, this approach plateaus early during training and yields a severe overall performance drop of 2.6 absolute points. This confirms that standard self-play without structural guidance is insufficient for complex tasks.

\textbf{Over-specialization.} Generating tasks exclusively from the skill library via the \emph{Skill-only pool} degrades generalized evaluation performance on API-Bank (Figure~\ref{fig:toolcall_data_efficiency}b). This validates our dual-stream architecture: mixing open-ended exploration is crucial to preserve task diversity and prevent mode collapse.

\textbf{Static skill orchestration.} Both \emph{Uniform routing} and \emph{Frozen skills} strictly underperform the full system, causing overall accuracy drops of 1.9 and 2.3 points respectively (Table~\ref{tab:ablations}). This proves that the performance gains stem from dynamic curriculum allocation and continuous library evolution rather than merely from injecting static structural constraints.

\input{tables/ablation}

\textbf{Co-evolutionary updates.} Table~\ref{tab:frozen_component_ablations_best} evaluates the necessity of continuous policy optimization. The \emph{Frozen proposer} variant fixes the generator at initialization, causing a 2.1-point overall drop. This shows that the proposer must learn to leverage the evolving skill library via RL. More critically, the \emph{Frozen feedback solver} variant fixes the solver used for probing and proposer rewards. This severs the dynamic frontier-tracking mechanism, supplying the proposer with outdated difficulty signals and leading to a severe 3.0-point degradation. Finally, \emph{Frozen both} disables the co-evolutionary loop, yielding the largest performance drop (-3.2 points). This confirms that proactive curriculum generation requires both policies to co-evolve: the proposer must adapt to synthesize harder tasks, while the feedback solver must advance to provide accurate boundary evaluations.

\input{tables/ablation_frozen_components}

\subsection{Data-Loop Diagnostics}
\label{sec:data_loop_diagnostics}

\begin{figure*}[htbp]
    \centering
    \begin{minipage}[t]{0.31\textwidth}
        \centering
        \includegraphics[width=\linewidth]{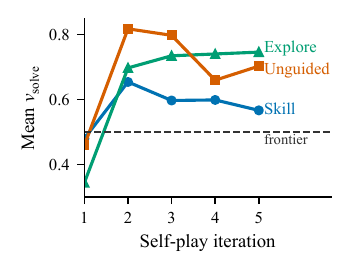}\\[-0.4em]
        {\small (a) Frontier proximity.}
    \end{minipage}
    \hfill
    \begin{minipage}[t]{0.31\textwidth}
        \centering
        \includegraphics[width=\linewidth]{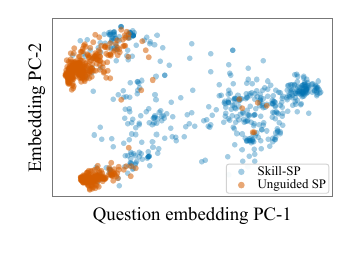}\\[-0.4em]
        {\small (b) Task diversity.}
    \end{minipage}
    \hfill
    \begin{minipage}[t]{0.31\textwidth}
        \centering
        \includegraphics[width=\linewidth]{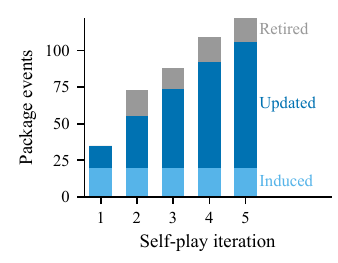}\\[-0.4em]
        {\small (c) Skill lifecycle.}
    \end{minipage}
    \caption{\textbf{Data-loop diagnostics for Qwen3-4B-Instruct tool-call self-play.}
    Skill-stream records stay closer to the empirical frontier than both the exploration stream and Unguided SP, while the overall Skill-SP pool yields broader task coverage in the question-embedding space, and the skill library is continually induced, updated, and retired across self-play iterations.}
    \label{fig:data_loop_frontier_quality}
\end{figure*}

Final accuracy alone obscures the underlying mechanism of self-improvement. To verify that Skill-SP constructs a controlled curriculum rather than merely acting as a passive data filter, we audit the accepted records and the skill-library lifecycle illustrated in Figure~\ref{fig:data_loop_frontier_quality}.

\textbf{Curriculum Quality.} Figure~\ref{fig:data_loop_frontier_quality}a tracks the empirical solver success rate $v_\text{solve}$. The skill-routed stream consistently generates tasks closer to the solver's learning frontier with a mean $v_\text{solve}$ of approximately 0.57. This distinctly outperforms the Unguided SP and exploration streams, which drift to 0.70 and 0.75 respectively. In Figure~\ref{fig:data_loop_frontier_quality}b, we encode the generated questions using all-MiniLM-L6-v2~\citep{reimers-2019-sentence-bert} and project their embeddings via PCA. The visualization reveals that Unguided SP tasks cluster into narrow semantic regions, whereas the Skill-SP mixed pool demonstrates broader coverage in the embedding space. This demonstrates that skill orchestration simultaneously pushes task difficulty and fosters structural diversity.

\textbf{Library Evolution.} This frontier-targeted and diverse curriculum is driven by a highly dynamic interface. Across five iterations, Figure~\ref{fig:data_loop_frontier_quality}c shows that Skill-SP continuously induces roughly 20 new packages per round, updates existing ones based on execution traces, and retires obsolete skills. These diagnostics show that the skill library functions as an adaptive curriculum engine that expands agent capabilities.

\begin{wrapfigure}{r}{0.33\columnwidth}
    \vspace{-1.0em}
    \centering
    \includegraphics[width=\linewidth]{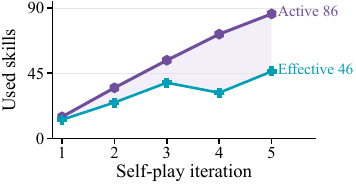}
    \vspace{-1.6em}
    \caption{\textbf{Skill utilization.} Active and effective skills grow steadily, driving broad curriculum expansion.}
    \label{fig:skill_utilization}
    \vspace{-1.0em}
\end{wrapfigure}

\textbf{Skill Utilization.} A growing library must actively diversify training to be useful. As shown in Figure~\ref{fig:skill_utilization}, the number of \emph{active} skills (yielding $\ge 1$ accepted record) expands to 86 across five iterations. To rigorously discount rare long-tail usage, we track \emph{effective} skills via exponentiated Shannon entropy, $N_{\mathrm{eff}} = \exp(-\sum_j p_j \log p_j)$ (where $p_j$ is the accepted frequency fraction), which steadily grows to 46. This continuous growth confirms that Skill-SP successfully converts an expanding library into a broad and balanced curriculum, ensuring the generator does not idle on a few dominant patterns.

%% file: tables/tool_call_main.tex
\definecolor{bggreen}{HTML}{E6F4EA}
\definecolor{bggray}{HTML}{F8F9FA}

\newcommand{\groupspace}{\noalign{\begingroup\color{white}\hrule height 0.5em\endgroup}}

\newcommand{\gainlast}[1]{\textsubscript{\textcolor{softgreen}{#1}}} 
\newcommand{\droplast}[1]{\textsubscript{\textcolor{softred}{#1}}} 
\newcommand{\gainphantomlast}[1]{\phantom{\textsubscript{#1}}}

\begin{table}[tbp] 
\centering
\footnotesize
\renewcommand{\arraystretch}{1.3} 

\definecolor{softgreen}{rgb}{0.12, 0.60, 0.12} 
\definecolor{softred}{rgb}{0.75, 0.25, 0.25} 

\newcommand{\gain}[1]{\rlap{\textsubscript{\textcolor{softgreen}{#1}}}} 
\newcommand{\drop}[1]{\rlap{\textsubscript{\textcolor{softred}{#1}}}} 
\newcommand{\same}[1]{\rlap{\textsubscript{\textcolor{black}{#1}}}}

\begin{NiceTabular*}{\textwidth}{l @{\extracolsep{\fill}} cccc ccccc c}[colortbl-like]
\toprule
\multirow{2}{*}{\textbf{Method}} & \multicolumn{4}{c}{\textbf{API-Bank}} & \multicolumn{5}{c}{\textbf{BFCL}} & \multirow{2}{*}{\textbf{Avg.}} \\
\cmidrule(lr){2-5} \cmidrule(lr){6-10}
 & L1 & L2 & L3 & Avg. & JS & Py & Java & Live & Avg. & \\
\midrule
\rowcolor{bggray} Qwen3-4B-Ins & 58.1 & 42.5 & 35.6 & 51.4 & 57.8 & 91.8 & 59.8 & 75.6 & 71.2 & 60.2\gainphantomlast{+6.5} \\
\rowcolor{bggray} \quad + Unguided SP & 60.2\gain{+2.1} & 46.3\gain{+3.8} & 40.8\gain{+5.2} & 54.4\gain{+3.0} & 65.8\gain{+8.0} & 95.0\gain{+3.2} & 63.0\gain{+3.2} & 77.9\gain{+2.3} & 75.4\gain{+4.2} & 64.1\gainlast{+3.9} \\
\rowcolor{bggreen} \quad \textbf{+ Skill-SP} & \textbf{64.6}\gain{+6.5} & \textbf{54.7}\gain{+12.2} & \textbf{44.0}\gain{+8.4} & \textbf{58.9}\gain{+7.5} & \textbf{65.5}\gain{+7.7} & \textbf{96.0}\gain{+4.2} & \textbf{63.5}\gain{+3.7} & \textbf{78.5}\gain{+2.9} & \textbf{75.9}\gain{+4.7} & \textbf{66.7}\gainlast{+6.5} \\
\groupspace

\rowcolor{bggray} Qwen3-8B & 73.3 & 61.4 & 43.9 & 65.5 & 69.0 & 95.1 & 64.9 & 77.9 & 76.7 & 69.4\gainphantomlast{+2.8} \\
\rowcolor{bggray} \quad + Unguided SP & 74.3\gain{+1.0} & 64.4\gain{+3.0} & 48.3\gain{+4.4} & 67.5\gain{+2.0} & 71.8\gain{+2.8} & 95.8\gain{+0.7} & 64.9\same{+0.0} & 77.8\drop{-0.1} & 77.5\gain{+0.8} & 71.0\gainlast{+1.6} \\
\rowcolor{bggreen} \quad \textbf{+ Skill-SP} & \textbf{75.9}\gain{+2.6} & \textbf{67.2}\gain{+5.8} & \textbf{50.0}\gain{+6.1} & \textbf{69.2}\gain{+3.7} & \textbf{72.8}\gain{+3.8} & \textbf{96.0}\gain{+0.9} & \textbf{65.1}\gain{+0.2} & \textbf{78.7}\gain{+0.8} & \textbf{78.2}\gain{+1.5} & \textbf{72.2}\gainlast{+2.8} \\
\groupspace

\rowcolor{bggray} Ministral-3-8B & 37.6 & 35.6 & 20.7 & 33.7 & 8.2 & 16.4 & 9.2 & 17.3 & 12.8 & 20.7\gainphantomlast{+42.9} \\
\rowcolor{bggray} \quad + Unguided SP & 38.5\gain{+0.9} & 36.4\gain{+0.8} & 20.8\gain{+0.1} & 34.4\gain{+0.7} & 7.5\drop{-0.7} & 15.4\drop{-1.0} & 8.5\drop{-0.7} & 18.8\gain{+1.5} & 12.5\drop{-0.3} & 20.8\gainlast{+0.1\phantom{9}} \\
\rowcolor{bggreen} \quad \textbf{+ Skill-SP} & \textbf{68.5}\gain{+30.9} & \textbf{56.2}\gain{+20.6} & \textbf{42.9}\gain{+22.2} & \textbf{61.5}\gain{+27.8} & \textbf{48.5}\gain{+40.3} & \textbf{90.8}\gain{+74.4} & \textbf{58.1}\gain{+48.9} & \textbf{80.0}\gain{+62.7} & \textbf{69.3}\gain{+56.5} & \textbf{63.6}\gainlast{+42.9} \\
\groupspace

\rowcolor{bggray} Ministral-3-14B & 31.1 & 15.5 & 15.7 & 26.0 & 14.5 & 32.9 & 18.6 & 27.2 & 23.3 & 22.2\gainphantomlast{+42.3} \\
\rowcolor{bggray} \quad + Unguided SP & 67.2\gain{+36.1} & 50.6\gain{+35.1} & 46.3\gain{+30.6} & 60.8\gain{+34.8} & 36.8\gain{+22.3} & 88.7\gain{+55.8} & 46.1\gain{+27.5} & 77.2\gain{+50.0} & 62.2\gain{+38.9} & 59.0\gainlast{+36.8} \\
\rowcolor{bggreen} \quad \textbf{+ Skill-SP} & \textbf{70.9}\gain{+39.8} & \textbf{57.5}\gain{+42.0} & \textbf{48.9}\gain{+33.2} & \textbf{64.5}\gain{+38.5} & \textbf{42.0}\gain{+27.5} & \textbf{89.9}\gain{+57.0} & \textbf{59.0}\gain{+40.4} & \textbf{83.1}\gain{+55.9} & \textbf{68.5}\gain{+45.2} & \textbf{64.5}\gainlast{+42.3} \\
\groupspace

\rowcolor{bggray} Granite-4.1-3B & 57.5 & 53.7 & 43.2 & 53.9 & 66.5 & 66.5 & 57.4 & 55.9 & 61.6 & 57.2\gainphantomlast{+5.3} \\
\rowcolor{bggray} \quad + Unguided SP & 59.0\gain{+1.5} & 54.3\gain{+0.6} & 42.9\drop{-0.3} & 54.9\gain{+1.0} & 63.5\drop{-3.0} & 70.6\gain{+4.1} & 57.9\gain{+0.5} & 59.5\gain{+3.6} & 62.9\gain{+1.3} & 58.2\gainlast{+1.0} \\
\rowcolor{bggreen} \quad \textbf{+ Skill-SP} & \textbf{63.2}\gain{+5.7} & \textbf{57.1}\gain{+3.4} & \textbf{47.0}\gain{+3.8} & \textbf{59.0}\gain{+5.1} & \textbf{67.5}\gain{+1.0} & \textbf{78.7}\gain{+12.2} & \textbf{57.9}\gain{+0.5} & \textbf{65.9}\gain{+10.0} & \textbf{67.5}\gain{+5.9} & \textbf{62.5}\gainlast{+5.3} \\
\bottomrule
\end{NiceTabular*}
\caption{\textbf{Main tool-call prediction results (avg@8).} Models are evaluated on API-Bank across three difficulty levels (L1--L3) and on BFCL across diverse programming languages and live scenarios. Absolute changes over base models are shown as subscripts: gains in green (\textcolor{softgreen}{\textbf{+Gain}}), drops in red (\textcolor{softred}{\textbf{-Drop}}).}
\label{tab:tool_call_main}
\end{table}

%% file: tables/logical_main.tex
\definecolor{bggreen}{HTML}{E6F4EA}

\begin{table}[tbp]
\centering
\footnotesize 
\renewcommand{\arraystretch}{1.3} 

\definecolor{softgreen}{rgb}{0.12, 0.60, 0.12} 
\providecommand{\gain}[1]{\rlap{\textsubscript{\textcolor{softgreen}{#1}}}} 
\newcommand{\same}[1]{\rlap{\textsubscript{\textcolor{black}{#1}}}}

\begin{NiceTabular*}{0.9\textwidth}{l @{\extracolsep{\fill}} cccccc @{\hspace{2em}}}[colortbl-like]
\toprule
\multirow{2}{*}{\textbf{Method}} & \textbf{Overall} & \textcolor[rgb]{0, 0.8, 0}{$\bullet$}~\textbf{Small} & \textcolor{blue}{$\blacksquare$}~\textbf{Medium} & $\blacklozenge$~\textbf{Large} & $\blacklozenge\blacklozenge$~\textbf{X-Large} & \textbf{Cell-level} \\
 & \scriptsize Grid-level acc. & \scriptsize $< 10^3$ & \scriptsize $10^3 \sim 10^6$ & \scriptsize $10^6 \sim 10^9$ & \scriptsize $> 10^9$ & \textbf{Acc.} \\
\midrule

\rowcolor{bggray} Qwen3-4B-Ins & 72.1 & 97.2 & 88.6 & 62.0 & 18.7 & 70.3 \\
\rowcolor{bggreen} \quad \textbf{+ Skill-SP} & \textbf{73.5}\gain{+1.4} & \textbf{97.3}\gain{+0.1} & \textbf{89.7}\gain{+1.1} & \textbf{64.6}\gain{+2.6} & \textbf{21.9}\gain{+3.2} & \textbf{74.2}\gain{+3.9} \\
\groupspace

\rowcolor{bggray} Qwen3-8B & 23.6 & 67.2 & 7.3 & 0.3 & 0.0 & 43.8 \\
\rowcolor{bggreen} \quad \textbf{+ Skill-SP} & \textbf{32.4}\gain{+8.8} & \textbf{82.0}\gain{+14.8} & \textbf{20.7}\gain{+13.4} & \textbf{1.8}\gain{+1.6} & \textbf{0.1}\gain{+0.1} & \textbf{49.1}\gain{+5.4} \\
\groupspace

\rowcolor{bggray} Ministral-3-8B & 5.0 & 14.1 & 1.6 & 0.0 & 0.0 & 3.7 \\
\rowcolor{bggreen} \quad \textbf{+ Skill-SP} & \textbf{11.2}\gain{+6.2} & \textbf{32.8}\gain{+18.7} & \textbf{2.5}\gain{+0.9} & \textbf{0.1}\gain{+0.1} & \textbf{0.0}\same{+0.0} & \textbf{23.7}\gain{+20.0} \\
\groupspace

\rowcolor{bggray} Ministral-3-14B & 5.4 & 14.9 & 2.2 & 0.2 & 0.0 & 7.3 \\
\rowcolor{bggreen} \quad \textbf{+ Skill-SP} & \textbf{17.4}\gain{+12.0} & \textbf{50.2}\gain{+35.3} & \textbf{4.6}\gain{+2.4} & \textbf{0.2}\same{+0.0} & \textbf{0.0}\same{+0.0} & \textbf{26.4}\gain{+19.1} \\
\groupspace

\rowcolor{bggray} Granite-4.1-3B & 11.6 & 35.4 & 0.8 & 0.1 & 0.0 & 34.3 \\
\rowcolor{bggreen} \quad \textbf{+ Skill-SP} & \textbf{12.6}\gain{+1.0} & \textbf{38.4}\gain{+3.0} & \textbf{1.2}\gain{+0.4} & \textbf{0.1}\same{+0.0} & \textbf{0.0}\same{+0.0} & \textbf{35.1}\gain{+0.8} \\
\bottomrule
\end{NiceTabular*}
\caption{\textbf{Main logical reasoning results (avg@8).} Models are evaluated on the ZebraLogic benchmark across Small, Medium, Large, and X-Large task scales, where scale is determined by search space size. Overall Grid-level accuracy and Cell-level accuracy are reported. Absolute gains of Skill-SP over the corresponding base models are shown as subscripts (\textcolor{softgreen}{\textbf{+Gain}}) next to the scores.}
\label{tab:logical_main}
\end{table}

%% file: tables/ablation.tex
\begin{table}[htbp]
\centering
\footnotesize 
\renewcommand{\arraystretch}{1.3} 

\definecolor{softgreen}{rgb}{0.12, 0.60, 0.12} 
\definecolor{softred}{rgb}{0.75, 0.25, 0.25} 

\newcommand{\gain}[1]{\rlap{\textsubscript{\textcolor{softgreen}{#1}}}} 
\newcommand{\drop}[1]{\rlap{\textsubscript{\textcolor{softred}{#1}}}} 
\newcommand{\same}[1]{\rlap{\textsubscript{\textcolor{black}{#1}}}}

\begin{tabular*}{1\textwidth}{l @{\extracolsep{\fill}} cccc ccccc c}
\toprule
\multirow{2}{*}{\textbf{Ablation Variant}} & \multicolumn{4}{c}{\textbf{API-Bank}} & \multicolumn{5}{c}{\textbf{BFCL}} & \multirow{2}{*}{\textbf{Overall}} \\
\cmidrule(lr){2-5} \cmidrule(lr){6-10}
 & L1 & L2 & L3 & Avg. & JS & Py & Java & Live & Avg. & \\
\midrule
\textbf{Full System} & \textbf{64.6} & \textbf{54.7} & \textbf{44.0} & \textbf{58.9} & \textbf{65.5} & \textbf{96.0} & \textbf{63.5} & \textbf{78.5} & \textbf{75.9} & \textbf{66.7} \\
\midrule 
\addlinespace[0.2em]

\textbf{Unguided SP} & 60.2\drop{-4.4} & 46.3\drop{-8.4} & 40.8\drop{-3.2} & 54.4\drop{-4.5} & 65.8\gain{+0.3} & 95.0\drop{-1.0} & 63.0\drop{-0.5} & 77.9\drop{-0.6} & 75.4\drop{-0.5} & 64.1\drop{-2.6} \\
\addlinespace[0.2em]

\textbf{Uniform routing} & 60.8\drop{-3.8} & 50.4\drop{-4.3} & 39.7\drop{-4.3} & 55.0\drop{-3.9} & 64.0\drop{-1.5} & 95.9\drop{-0.1} & 64.1\gain{+0.6} & 78.5\same{+0.0} & 75.6\drop{-0.3} & 64.8\drop{-1.9} \\
\addlinespace[0.2em]

\textbf{Frozen skills} & 61.3\drop{-3.3} & 50.2\drop{-4.5} & 40.2\drop{-3.8} & 55.4\drop{-3.5} & 61.5\drop{-4.0} & 95.5\drop{-0.5} & 63.1\drop{-0.4} & 78.9\gain{+0.4} & 74.8\drop{-1.1} & 64.4\drop{-2.3} \\
\bottomrule
\end{tabular*}
\vspace{1mm}
\caption{\textbf{Quantitative ablations on the self-play data loop.} All variants are initialized with the Qwen3-4B-Instruct backbone, with each disabling a specific component of the proposal-side orchestration. Subscripts (\textcolor{softred}{\textbf{-Drop}}) indicate the absolute performance degradation compared to the full system.}
\label{tab:ablations}
\end{table}

%% file: tables/ablation_frozen_components.tex
\begin{table}[htbp]
\centering
\footnotesize
\renewcommand{\arraystretch}{1.3}

\definecolor{frozenbestred}{rgb}{0.75, 0.25, 0.25}
\newcommand{\frozenbestdrop}[1]{\rlap{\textsubscript{\textcolor{frozenbestred}{#1}}}}
\newcommand{\frozenbestsame}[1]{\rlap{\textsubscript{\textcolor{black}{#1}}}}

\begin{tabular*}{1\textwidth}{l @{\extracolsep{\fill}} cccc ccccc c}
\toprule
\multirow{2}{*}{\textbf{Ablation Variant}} & \multicolumn{4}{c}{\textbf{API-Bank}} & \multicolumn{5}{c}{\textbf{BFCL}} & \multirow{2}{*}{\textbf{Overall}} \\
\cmidrule(lr){2-5} \cmidrule(lr){6-10}
 & L1 & L2 & L3 & Avg. & JS & Py & Java & Live & Avg. & \\
\midrule
\textbf{Full System} & \textbf{64.6} & \textbf{54.7} & \textbf{44.0} & \textbf{58.9} & \textbf{65.5} & \textbf{96.0} & \textbf{63.5} & \textbf{78.5} & \textbf{75.9} & \textbf{66.7} \\
\midrule
\addlinespace[0.2em]

\textbf{Frozen proposer} & 62.0\frozenbestdrop{-2.6} & 53.0\frozenbestdrop{-1.7} & 43.8\frozenbestdrop{-0.2} & 57.0\frozenbestdrop{-1.9} & 55.5\frozenbestdrop{-10.0} & 94.0\frozenbestdrop{-2.0} & 62.6\frozenbestdrop{-0.9} & 78.1\frozenbestdrop{-0.4} & 72.6\frozenbestdrop{-3.3} & 64.6\frozenbestdrop{-2.1} \\
\addlinespace[0.2em]

\textbf{Frozen feedback solver} & 59.9\frozenbestdrop{-4.7} & 45.2\frozenbestdrop{-9.5} & 39.3\frozenbestdrop{-4.7} & 53.7\frozenbestdrop{-5.2} & 65.3\frozenbestdrop{-0.3} & 95.5\frozenbestdrop{-0.5} & 62.8\frozenbestdrop{-0.8} & 78.2\frozenbestdrop{-0.3} & 75.4\frozenbestdrop{-0.5} & 63.7\frozenbestdrop{-3.0} \\
\addlinespace[0.2em]

\textbf{Frozen both} & 59.2\frozenbestdrop{-5.4} & 47.6\frozenbestdrop{-7.1} & 39.2\frozenbestdrop{-4.8} & 53.5\frozenbestdrop{-5.4} & 63.5\frozenbestdrop{-2.0} & 94.3\frozenbestdrop{-1.8} & 62.0\frozenbestdrop{-1.5} & 78.5\frozenbestsame{+0.0} & 74.6\frozenbestdrop{-1.3} & 63.5\frozenbestdrop{-3.2} \\
\bottomrule
\end{tabular*}
\vspace{1mm}
\caption{\textbf{Quantitative ablations on the co-evolutionary update loop.} All variants are initialized with the Qwen3-4B-Instruct backbone, with each disabling a specific component of the co-evolutionary update loop. Subscripts (\textcolor{frozenbestred}{\textbf{-Drop}}) indicate the absolute performance degradation compared to the full system.}
\label{tab:frozen_component_ablations_best}
\end{table}

%% file: Sections/6_Conclusion.tex
\section{Conclusion}
\label{sec:conclusion}

We presented \emph{\textbf{Skill} \textbf{S}elf-\textbf{P}lay} (\textbf{Skill-SP}), a co-evolutionary framework designed to resolve the fundamental tension between task diversity and verification reliability in LLM self-evolution. 
By leveraging a dynamically evolving library of modular skills as proactive task-pattern interfaces, Skill-SP successfully steers the proposer to synthesize high-fidelity curricula tailored precisely to the solver's current learning frontier. 
This structured abstraction effectively mitigates the context bottlenecks of raw trajectory histories, while equipping the self-play loop with explicit, executable boundaries for rigorous verification. 
Our extensive evaluations across complex tool-calling and logical reasoning domains confirm that this dynamic orchestration systematically expands the performance horizons of diverse language models. Ultimately, our findings demonstrate that co-evolving skills serve as powerful training-time scaffolds, unlocking a sustainable, structured, and truly open-ended pathway for the autonomous self-evolution of large language models.

%% file: Sections/X_Appendix.tex
\section{Iteration-wise Evaluation Trajectories}
\label{app:iteration_trajectories}

\begin{figure*}[htbp]
    \centering
    \begin{minipage}[t]{0.37\textwidth}
        \centering
        \includegraphics[width=\linewidth]{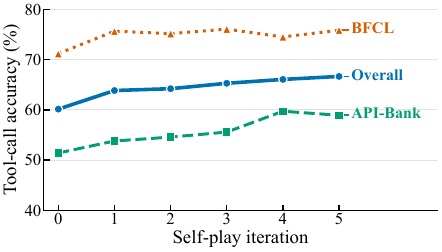}\\
        \small \textbf{(a) Tool-call}
    \end{minipage}\hspace{0.03\textwidth}
    \begin{minipage}[t]{0.37\textwidth}
        \centering
        \includegraphics[width=\linewidth]{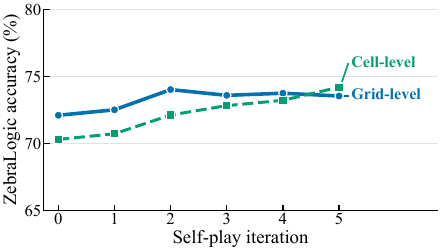}\\
        \small \textbf{(b) Logical reasoning}
    \end{minipage}
    \caption{\textbf{Qwen3-4B-Instruct.} Held-out tool-call (left) and logical-reasoning (right) trajectories.}
    \label{fig:app_iteration_qwen3_4b}
\end{figure*}

\begin{figure*}[htbp]
    \centering
    \begin{minipage}[t]{0.37\textwidth}
        \centering
        \includegraphics[width=\linewidth]{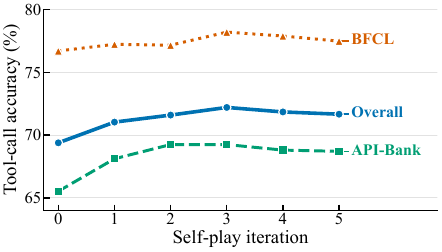}\\
        \small \textbf{(a) Tool-call}
    \end{minipage}\hspace{0.03\textwidth}
    \begin{minipage}[t]{0.37\textwidth}
        \centering
        \includegraphics[width=\linewidth]{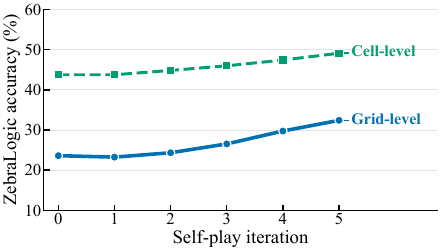}\\
        \small \textbf{(b) Logical reasoning}
    \end{minipage}
    \caption{\textbf{Qwen3-8B.} Held-out tool-call (left) and logical-reasoning (right) trajectories.}
    \label{fig:app_iteration_qwen3_8b}
\end{figure*}

\begin{figure*}[htbp]
    \centering
    \begin{minipage}[t]{0.37\textwidth}
        \centering
        \includegraphics[width=\linewidth]{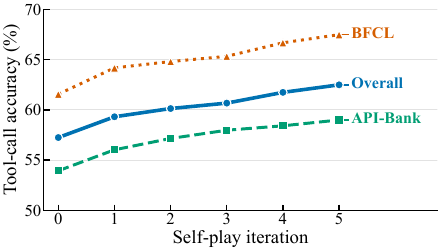}\\
        \small \textbf{(a) Tool-call}
    \end{minipage}\hspace{0.03\textwidth}
    \begin{minipage}[t]{0.37\textwidth}
        \centering
        \includegraphics[width=\linewidth]{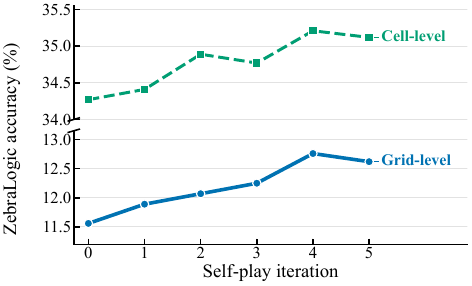}\\
        \small \textbf{(b) Logical reasoning}
    \end{minipage}
    \caption{\textbf{Granite-4.1-3B.} Held-out tool-call (left) and logical-reasoning (right) trajectories.}
    \label{fig:app_iteration_granite_3b}
\end{figure*}

\begin{figure*}[htbp]
    \centering
    \begin{minipage}[t]{0.37\textwidth}
        \centering
        \includegraphics[width=\linewidth]{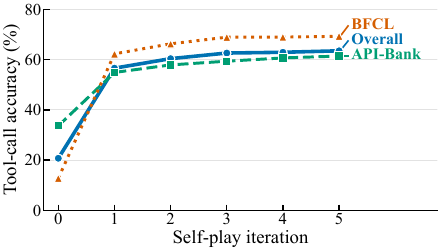}\\
        \small \textbf{(a) Tool-call}
    \end{minipage}\hspace{0.03\textwidth}
    \begin{minipage}[t]{0.37\textwidth}
        \centering
        \includegraphics[width=\linewidth]{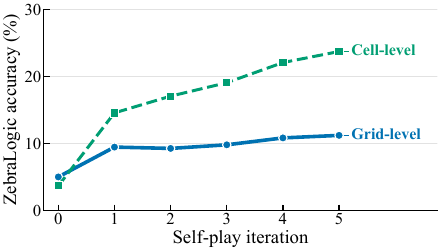}\\
        \small \textbf{(b) Logical reasoning}
    \end{minipage}
    \caption{\textbf{Ministral-3-8B.} Held-out tool-call (left) and logical-reasoning (right) trajectories.}
    \label{fig:app_iteration_ministral_8b}
\end{figure*}

\begin{figure*}[htbp]
    \centering
    \begin{minipage}[t]{0.37\textwidth}
        \centering
        \includegraphics[width=\linewidth]{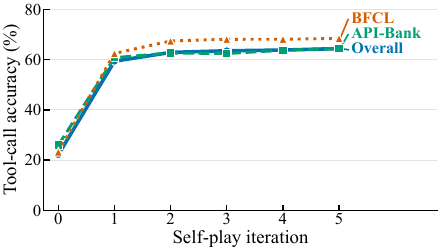}\\
        \small \textbf{(a) Tool-call}
    \end{minipage}\hspace{0.03\textwidth}
    \begin{minipage}[t]{0.37\textwidth}
        \centering
        \includegraphics[width=\linewidth]{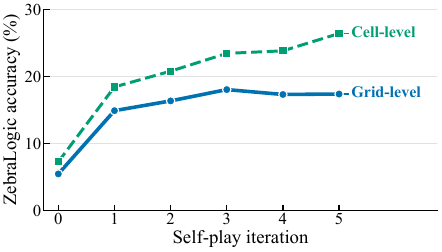}\\
        \small \textbf{(b) Logical reasoning}
    \end{minipage}
    \caption{\textbf{Ministral-3-14B.} Held-out tool-call (left) and logical-reasoning (right) trajectories.}
    \label{fig:app_iteration_ministral_14b}
\end{figure*}

Figures~\ref{fig:app_iteration_qwen3_4b}--\ref{fig:app_iteration_ministral_14b} report raw held-out performance after each self-play iteration, with iteration~0 denoting the initial solver. Tool-call panels track the seven-subset overall score, API-Bank, and BFCL; logical-reasoning panels track ZebraLogic grid-level and cell-level accuracy.

\section{Details of Dynamic Skill Routing}
\label{app:skill_routing}

To ensure that the proposer efficiently focuses on structural priors that yield high-quality tasks, Skill-SP routes each generation attempt by dynamically sampling a skill $\vs \sim \gS$. This routing mechanism must carefully balance exploiting historically reliable skills with exploring newly induced or under-tested ones.

For each skill $\vs_j \in \gS$, its tracking statistics $\sigma_j$ record the total number of historical selection attempts $a_j$. It also maintains the specific success counts across three crucial stages of validity verification (detailed in Section~\ref{sec:proposer_optimization}): the number of structurally verified candidates $n_j^{\mathrm{ver}}$, solver-consistent candidates $n_j^{\mathrm{con}}$, and frontier-difficulty tasks $n_j^{\mathrm{fd}}$. 

To mitigate the high variance in early sampling estimates, we apply symmetric Beta smoothing with a prior strength $\kappa$. The smoothed success rate for each outcome $o \in \{\mathrm{ver}, \mathrm{con}, \mathrm{fd}\}$ is calculated as:
\begin{equation}
    \widehat{r}^{o}_j = \frac{n^{o}_j + \kappa/2}{a_j + \kappa}.
\end{equation}
We then compute a composite quality score $v_\text{skill}(\vs_j)$ by taking a weighted average of these smoothed rates, prioritizing the fundamental structural validity while equally rewarding consistency and frontier difficulty:
\begin{equation}
    v_\text{skill}(\vs_j)
    =
    \frac{1}{2}\widehat{r}^{\mathrm{ver}}_j
    +
    \frac{1}{4}\widehat{r}^{\mathrm{con}}_j
    +
    \frac{1}{4}\widehat{r}^{\mathrm{fd}}_j.
    \label{eq:app_skill_quality}
\end{equation}

Finally, to prevent the routing distribution from prematurely collapsing onto a few dominant skills, we augment the sampling weight with a decaying exploration bonus. The sampling weight $w(\vs_j)$ and the resulting selection probability $P(\vs_j)$ are formulated as:
\begin{equation}
    w(\vs_j)
    =
    \mathrm{clip}\!\left(v_\text{skill}(\vs_j), w_{\min}, w_{\max}\right) \left(1+\beta\exp(-a_j/\tau)\right),
    \qquad
    P(\vs_j)
    =
    \frac{w(\vs_j)}{\sum_{\vs_{\ell}\in\gS} w(\vs_{\ell})},
    \label{eq:app_skill_sampling}
\end{equation}
where the clipping bounds $[w_{\min}, w_{\max}]$ ensure numerical stability, $\beta$ controls the initial magnitude of the exploration bonus, and $\tau$ dictates its decay rate. This mechanism guarantees that under-tested skills are sampled sufficiently until their true empirical yield converges.

\section{Training and Implementation Details}
\label{app:training_details}

This section reports the settings that materially affect the generated curriculum, solver optimization, and final benchmark scores. Each training run uses eight NVIDIA A800 GPUs.

\paragraph{Task rewards.}
For tool-call prediction, the solver reward combines exact tool-call correctness and output format compliance:
\begin{equation}
    \mathcal{R}_{\mathrm{tool}}(\vx,\vy,\vc)
    =
    0.9 \times \mathbf{1}\{\mathrm{ExactCall}(\vy,\vc)\}
    +
    0.1 \times \mathbf{1}\{\mathrm{ValidFormat}(\vy)\}.
    \label{eq:tool_reward}
\end{equation}
Here, $\mathrm{ValidFormat}(\vy)$ indicates that the output contains at least one parseable tool call, and $\mathrm{ExactCall}(\vy,\vc)$ indicates that at least one parsed call exactly matches the reference function name and JSON arguments in the hidden contract $\vc$.
For logical reasoning, the solver reward is a sparse puzzle-level signal:
\begin{equation}
    \mathcal{R}_{\mathrm{logic}}(\vx,\vy,\vc)
    =
    0.9 \times \mathbf{1}\{\mathrm{FullPuzzle}(\vy,\vc)\}
    +
    0.1 \times \mathbf{1}\{\mathrm{ValidStructure}(\vy,\vc)\}.
    \label{eq:logic_reward}
\end{equation}
$\mathrm{ValidStructure}(\vy,\vc)$ requires the output to parse into a legal Zebra grid with the correct house and attribute schema, while $\mathrm{FullPuzzle}(\vy,\vc)$ requires all cells to match the hidden solution.
Cell accuracy is logged as a diagnostic metric but is not used in the training reward.

\paragraph{Proposer repetition penalty.}
To encourage generation diversity and prevent batch-level mode collapse, we subtract a repetition penalty $\rho(\vx)$ from the proposer's RL reward during optimization. Within a proposer rollout batch of $B$ candidates, the penalty for a generated prompt $\vx_i$ is computed as $\rho(\vx_i)=|C_i|/B$, where $C_i$ is the greedy cluster of task texts formed using a Jaccard similarity threshold of $0.5$.

For tool-call prediction, the frontier filter is applied after both structural parsing and solver probing.
Skill-stream records additionally use package-level validators, while exploration-stream records rely on global schema validity and solver consistency before entering $\mathcal{D}^{(t)}$.
For logical reasoning, the deterministic puzzle checker provides a stronger structural gate: generated puzzles must be parseable, valid, and uniquely solvable before they are used for solver training.
Across both domains, the final solver never receives skill-package context at training or evaluation time; it only observes the standard prompt $\vx$.

\begin{table}[htbp]
    \centering
    \small
    \begin{tabular}{p{0.22\linewidth} p{0.34\linewidth} p{0.34\linewidth}}
        \toprule
        \textbf{Setting} & \textbf{Tool-call prediction} & \textbf{Logical reasoning} \\
        \midrule
        Self-play loop & $T=5$ iterations & $T=5$ iterations \\
        \addlinespace
        Proposer update & GRPO with four rollouts per prompt and five update steps per iteration & GRPO with four rollouts per prompt and three update steps per iteration \\
        \addlinespace
        Solver update & Five rollouts per task group and 15 GRPO steps per iteration & Five rollouts per task group and 10 GRPO steps per iteration \\
        \addlinespace
        Training pool & $M=8{,}000$ accepted tasks per iteration with $\alpha=0.5$ skill-stream records and $0.5$ exploration-stream records & $M=1{,}920$ checker-verified puzzles per iteration \\
        \addlinespace
        Candidate filtering & Probe each candidate with $K=10$ frozen-solver responses; retain records with $v_\text{solve}(\vx,\vc;\pi_\text{solve}^{(t)})\in[0.25,0.75]$, solver consistency, and a unique majority answer & Enforce parser validity, puzzle validity, and uniqueness through the deterministic puzzle checker before solver training \\
        \addlinespace
        Skill routing & $\kappa=4$; $[w_{\min},w_{\max}]=[0.25,1.75]$; $\beta=1$, $\tau=8$; quality and exploration weighting enabled, age decay disabled & Same as tool-call prediction \\
        \addlinespace
        Optimization & AdamW with learning rate $10^{-6}$, weight decay $10^{-2}$, gradient clipping at $1.0$, and KL coefficient $10^{-2}$ & Same as tool-call prediction \\
        \addlinespace
        Context and generation & 8192-token prompt and response limits by default; the 14B Ministral tool-call runs use a 16384-token prompt limit & 8192-token prompt and response limits \\
        \addlinespace
        Skill evolution & Induce at least 20 new packages from accepted exploration records and refine packages after each iteration & Induce at least 10 new packages from verified puzzle records and refine packages after each iteration \\
        \addlinespace
        Benchmark evaluation & avg@8 sampling with temperature $1.0$, top-$p=1.0$, and top-$k=40$ & avg@8 sampling with temperature $1.0$, top-$p=1.0$, and top-$k=40$ \\
        \bottomrule
    \end{tabular}
    \caption{\textbf{Training and evaluation details.} We report the main settings used for both task families.}
    \label{tab:training_details}
\end{table}

\section{Initial Skill Construction}
\label{app:initial_skills}

\paragraph{Construction.}
We construct each task family's initial skill library $\gS^{(0)}$ once and share it across all model backbones. The reported libraries contain 15 generic tool-call packages and 8 generic ZebraLogic packages. An LLM proposes these packages from domain-level task formats and verifier interfaces using the same representation as Skill-SP, $\vs=\langle m,r,h,e,\nu,\sigma\rangle$. We retain only packages whose required fields can be parsed and whose domain-specific components pass structural and verification checks.

\paragraph{Effect of online evolution.}
The \emph{Frozen skills} ablation in Table~\ref{tab:ablations} isolates the contribution of this shared initialization by holding $\gS^{(0)}$ fixed while retaining skill-guided generation. Frozen skills improves only marginally over Unguided SP ($64.4$ vs. $64.1$ overall) and remains below the full system ($66.7$). Thus, most of the gain arises from online skill evolution rather than the initial library alone.

\section{Full Pseudocode for Skill-SP}
\label{app:osp_pseudocode}

Algorithm~\ref{alg:skill_sp} expands the high-level workflow in Section~\ref{sec:methodology} into a single training loop.

\begin{algorithm}[htbp]
    \caption{\textsc{Skill Self-Play} (Skill-SP)}
    \label{alg:skill_sp}
    \KwIn{Initial proposer $\pi_\text{propose}^{(0)}$, solver $\pi_\text{solve}^{(0)}$, and skill library $\gS^{(0)}$; self-play iterations $T$; target pool size $M$; curriculum ratio $\alpha$; solver probe count $K$.}
    \KwOut{Final solver $\pi_\text{solve}^{(T)}$.}

    \For{$t = 0,1,\ldots,T-1$}{
        Initialize $\mathcal{T}_{\text{skill}}^{(t)} \gets \emptyset$ and $\mathcal{T}_{\text{explore}}^{(t)} \gets \emptyset$\;

        \ForEach{skill-conditioned proposal attempt}{
            Sample $\vs \sim \gS^{(t)}$ using its routing statistics $\sigma$\;
            Generate $(\vx,\vc) \sim \pi_\text{propose}^{(t)}(\cdot \mid \vs)$\;
            Draw $K$ responses from $\pi_\text{solve}^{(t)}(\cdot \mid \vx)$ to evaluate $v_\text{solve}(\vx,\vc;\pi_\text{solve}^{(t)})$, $\mathbb{1}_{\{(\vx,\vc) \text{ is valid}\}}$, and $\gR_\text{propose}(\vx,\vc;\pi_\text{solve}^{(t)})$\;
            \If{$\mathbb{1}_{\{(\vx,\vc) \text{ is valid}\}}=1$}{
                Add $(\vx,\vc,\vs)$ to $\mathcal{T}_{\text{skill}}^{(t)}$\;
            }
        }

        \ForEach{open-ended proposal attempt}{
            Generate $(\vx,\vc) \sim \pi_\text{propose}^{(t)}(\cdot \mid \varnothing)$\;
            Draw $K$ responses from $\pi_\text{solve}^{(t)}(\cdot \mid \vx)$ to evaluate $v_\text{solve}(\vx,\vc;\pi_\text{solve}^{(t)})$, $\mathbb{1}_{\{(\vx,\vc) \text{ is valid}\}}$, and $\gR_\text{propose}(\vx,\vc;\pi_\text{solve}^{(t)})$\;
            \If{$\mathbb{1}_{\{(\vx,\vc) \text{ is valid}\}}=1$}{
                Add $(\vx,\vc)$ to $\mathcal{T}_{\text{explore}}^{(t)}$\;
            }
        }

        Construct $\mathcal{D}^{(t)}$ from $\mathcal{T}_{\text{skill}}^{(t)}$ and $\mathcal{T}_{\text{explore}}^{(t)}$ using Eq.~\ref{eq:mixed_pool}\;
        Update $\pi_\text{propose}^{(t)} \rightarrow \pi_\text{propose}^{(t+1)}$ with GRPO according to Eq.~\ref{eq:proposer_rl_objective}, keeping $\pi_\text{solve}^{(t)}$ fixed\;
        Update $\pi_\text{solve}^{(t)} \rightarrow \pi_\text{solve}^{(t+1)}$ with GRPO on $\mathcal{D}^{(t)}$ according to Eq.~\ref{eq:solver_rl_objective}\;
        Update $\sigma$ from skill-conditioned attempts; refine skills from invalid attempts and accepted execution traces\;
        Build $\mathcal{T}^{(t)}_{\text{induce}}$ by retaining frontier-targeted samples from $\mathcal{T}_{\text{explore}}^{(t)}$\;
        Induce $\gS_{\text{new}}^{(t)}$ from $\mathcal{T}^{(t)}_{\text{induce}}$ and identify $\mathcal{S}_{\text{prune}}^{(t)}$\;
        Update the active library as $\gS^{(t+1)} \gets (\gS^{(t)} \setminus \mathcal{S}_{\text{prune}}^{(t)}) \cup \gS_{\text{new}}^{(t)}$\;
    }

    \Return{$\pi_\text{solve}^{(T)}$}\;
\end{algorithm}

\section{Computational Overhead}
\label{app:computational_overhead}

A complete five-iteration Skill-SP run takes slightly over one day on average in our setup. Table~\ref{tab:computational_overhead} reports the average wall-clock breakdown across complete runs. Proposer policy optimization includes its on-policy rollouts, reward computation, and GRPO update, whereas solver-curriculum construction is the subsequent collection pass that generates, probes, verifies, and mixes the training pool $\mathcal{D}^{(t)}$. Explicit skill induction and refinement together account for only $6.5\%$ of end-to-end runtime, indicating that evolving the skill library introduces limited computational overhead.

\begin{table}[htbp]
    \centering
    \small
    \begin{tabular}{l c}
        \toprule
        \textbf{Pipeline component} & \textbf{Runtime share} \\
        \midrule
        Proposer policy optimization & $24.3\%$ \\
        Solver-curriculum construction & $57.0\%$ \\
        \quad Skill-stream generation and verification & $20.0\%$ \\
        \quad Exploration-stream generation and verification & $37.0\%$ \\
        Solver policy optimization & $10.2\%$ \\
        \textbf{Skill-library evolution} & \textbf{$6.5\%$} \\
        \quad Skill induction & $1.3\%$ \\
        \quad Skill refinement & $5.2\%$ \\
        Held-out benchmark evaluation & $1.9\%$ \\
        Other orchestration & $0.1\%$ \\
        \bottomrule
    \end{tabular}
    \caption{\textbf{Average wall-clock breakdown.} Indented rows decompose their parent stage. Skill-library evolution occupies only a small fraction of the complete pipeline.}
    \label{tab:computational_overhead}
\end{table}

\section{Analysis of Evaluation Benchmark}
\label{app:evaluation_benchmark}

Our benchmark suite contains two complementary tool-call benchmarks, API-Bank and BFCL, and one logical-reasoning benchmark, ZebraLogic.
Together, they cover two forms of verifiable agent behavior: selecting and grounding executable actions from dialogue context, and solving deterministic constraint-satisfaction problems with a unique answer.

\begin{table}[htbp]
    \centering
    \small
    \begin{tabular}{>{\raggedright\arraybackslash}p{0.18\linewidth} >{\raggedright\arraybackslash}p{0.28\linewidth} >{\raggedright\arraybackslash}p{0.23\linewidth} >{\raggedright\arraybackslash}p{0.22\linewidth}}
        \toprule
        \textbf{Benchmark} & \textbf{Task Form} & \textbf{Main Capability} & \textbf{Reported Metrics} \\
        \midrule
        API-Bank \newline \citep{li2023api} & Dialogue-conditioned next tool call & Tool selection, argument grounding, schema adherence & Level 1--3 accuracy and average \\
        \addlinespace
        BFCL \newline \citep{patil2025berkeley} & Function-calling prompts across programming and live-tool categories & Cross-benchmark tool-call generalization & Category accuracy and BFCL average \\
        \addlinespace
        ZebraLogic \newline \citep{lin2025zebralogic} & Zebra-style grid puzzles with deterministic constraints & Constraint tracking, uniqueness reasoning, complete solution synthesis & Grid-level and cell-level accuracy \\
        \bottomrule
    \end{tabular}
    \caption{\textbf{Evaluation benchmark roles.} API-Bank and BFCL evaluate tool-call prediction from different benchmark sources, while ZebraLogic evaluates a separate logical-reasoning task family with deterministic verification.}
    \label{tab:benchmark_analysis}
\end{table}

\paragraph{API-Bank.}
API-Bank~\citep{li2023api} is the primary benchmark for evaluating tool-call prediction in our experiments.
Each instance is normalized into a standard record containing a system prompt with tool descriptions, a user-side dialogue history, and a hidden reference answer specifying the correct tool name and arguments.
The solver receives only the system and user messages and must output a single tool invocation.
Evaluation parses the predicted \texttt{<tool\_call>} block and exact-matches the normalized tool name and parameters against the reference answer.
We report accuracy on Levels 1--3, which lets us separate easier tool-routing cases from harder dialogue states that require more careful argument propagation and context grounding.
This benchmark therefore measures whether self-play improves the core operational skill that Skill-SP trains for: producing executable, schema-valid actions from standard prompts.

\paragraph{BFCL.}
BFCL~\citep{patil2025berkeley} complements API-Bank by evaluating tool-call generalization on a distinct benchmark source.
In the main paper, we report the displayed BFCL categories used in the seven-subset tool-call average: \textit{bfcl\_simple\_javascript}, \textit{bfcl\_simple\_python}, \textit{bfcl\_simple\_java}, and \textit{bfcl\_live\_simple}.
These categories stress different function-schema conventions, type systems, and argument normalization requirements.
Unlike API-Bank, BFCL is not merely a level-wise slice of the same benchmark family; it tests whether the solver's learned tool-use behavior transfers to new function descriptions and answer formats.
We use the benchmark-specific scorer to compare predicted calls against the provided possible answers after category-aware normalization.
BFCL therefore serves as a transfer check: improvements should persist when the solver faces different function descriptions, category conventions, and answer-normalization rules.

\paragraph{ZebraLogic.}
ZebraLogic~\citep{lin2025zebralogic} evaluates a different verifiable agent task family: logical reasoning over grid-structured constraint puzzles.
Each instance provides a natural-language puzzle whose solution is a complete assignment of attributes to houses.
The evaluator checks whether the model's output can be parsed into a valid grid, whether the predicted schema matches the reference schema, and whether each cell agrees with the hidden solution.
We report both grid-level accuracy, which requires the entire puzzle to be solved correctly, and cell-level accuracy, which measures partial progress.
The benchmark is further stratified by search-space scale into Small, Medium, Large, and X-Large subsets.
This stratification is useful for Skill-SP because it separates superficial format compliance from genuine constraint tracking: gains on larger or full-puzzle metrics indicate that the generated curriculum helps the solver learn harder reasoning patterns rather than only improving local answer formatting.
ZebraLogic also exposes the limitation of unguided self-play most clearly, since generating valid, uniquely solvable puzzles requires proactive structural guidance rather than passive post-hoc filtering.

\section{Limitations and Future Work}
\label{app:limitations}

While Skill-SP is highly effective for autonomous agent self-evolution, discovering entirely novel task patterns inherently requires the base model to possess a minimum foundational capability to bootstrap valid learning signals. For extremely complex domains, the framework might initially benefit from a small set of human demonstrations to jumpstart the evolving library. Furthermore, the current instantiation relies on fixed heuristics, such as the static skill-stream mixing ratio $\alpha$ and pre-defined difficulty bounds, which may require empirical tuning for new task families.

Building on these observations, future work will explore several promising directions. Replacing fixed routing heuristics with learnable, dynamic curriculum schedulers could autonomously optimize the data orchestration process. Alongside this, we aim to investigate the fully automated co-induction of generative rules and executable validators directly from raw environment interactions. Broadening the scope of the framework, transferring evolved skill libraries across different model architectures could enable strong frontier models to bootstrap smaller, efficient agents, presenting an exciting avenue for scalable and democratized alignment.

\section{Case Studies of Induced Skills}
\label{app:skill_case_studies}

All skill packages shown in this section were induced during Skill-SP runs initialized from Qwen3-4B-Instruct.

\subsection{Tool-Call Prediction}
\label{app:toolcall_skill_cases}

Each package records a reusable transition from an observed dialogue state to one valid next tool call, together with generation constraints and a local validator. The following examples illustrate library-level pattern induction.

Each complete listing is organized into routing metadata and curriculum state $(m,\sigma)$, proposer-visible construction fields $(r,h,e)$, and the package-local validator $\nu$. In the colored listings, blue denotes metadata and curriculum state, orange denotes procedural construction resources, green denotes hints and validation, and gray denotes a package example. Any listed solver procedure is an offline diagnostic oracle used to inspect generated samples.

Table~\ref{tab:toolcall_skill_cases} presents two packages whose metadata records \texttt{source=skill\_induction}. Both were induced in iteration~2 and specialize different forms of observation grounding. \texttt{skill\_038} turns a previously identified top-rated restaurant into the \texttt{restaurant\_id} of a reservation, while \texttt{skill\_042} uses a confirmed availability observation, rather than the user's broader time window, to construct a meeting interval.

\begin{table*}[htbp]
    \centering
    \small
    \renewcommand{\arraystretch}{1.22}
    \setlength{\tabcolsep}{3.5pt}
    \begin{tabular*}{\textwidth}{@{\extracolsep{\fill}} >{\raggedright\arraybackslash}p{0.16\textwidth} >{\raggedright\arraybackslash}p{0.23\textwidth} >{\raggedright\arraybackslash}p{0.29\textwidth} >{\raggedright\arraybackslash}p{0.25\textwidth}}
        \toprule
        \textbf{Induced package} & \textbf{Observed evidence} & \textbf{Induced grounding rule} & \textbf{Local validation and action} \\
        \midrule
        \texttt{skill\_038}\newline
        \emph{Discovery $\rightarrow$ reservation}\newline
        Iteration 2\newline
        $74/82$ filtered/routed
        & The user asks to reserve for four at 7 PM. A prior observation identifies the top-rated Italian restaurant and its identifier, \texttt{TRT-8842}.
        & Select \texttt{book\_table} after a rated discovery; copy \texttt{restaurant\_id} from the observation and normalize the requested date and time.
        & Require the observed identifier, complete arguments, ISO date/time, and a positive guest count.\newline
        \texttt{book\_table}: \texttt{TRT-8842}, \texttt{2023-09-16}, \texttt{19:00}, $4$ guests \\
        \addlinespace[0.45em]
        \texttt{skill\_042}\newline
        \emph{Availability $\rightarrow$ scheduling}\newline
        Iteration 2\newline
        $1{,}102/1{,}993$ filtered/routed
        & The user requests a meeting for Sarah and Alex between 10 AM and 2 PM, on ``project updates.'' A prior observation reports that only 10:30--11:30 is available.
        & Select \texttt{ScheduleMeeting} after the availability check; use the confirmed slot rather than the requested window, and copy the participants and topic.
        & Require ISO timestamps inside the requested window and evidence-grounded participants and topic.\newline
        \texttt{ScheduleMeeting}: Sarah, Alex; \texttt{10:30--11:30}; project updates \\
        \bottomrule
    \end{tabular*}
    \caption{\textbf{Tool-call case studies from the induced skill library.} All packages were created from self-play trajectories and saved with \texttt{source=skill\_induction}. Each package binds prior evidence to a next-action rule and checks that the resulting call remains grounded and schema-valid. The final line in each row compactly renders the verified call from its package example; $\mathrm{filtered}/\mathrm{routed}$ reports the package statistics accumulated during routing.}
    \label{tab:toolcall_skill_cases}
\end{table*}

\subsubsection{Complete induced package: \texttt{skill\_038}}
\label{app:complete_toolcall_skill}

The complete listing of \texttt{skill\_038} is organized into four color-coded blocks so that each package component remains readable at paper scale. Together, they reproduce all package-specific fields and one complete proposer example.

\newcommand{\skillcaseband}[2]{%
    \begingroup
    \setlength{\fboxsep}{4.5pt}%
    \noindent\colorbox[rgb]{#1}{\parbox{\dimexpr\linewidth-2\fboxsep\relax}{\small\textbf{#2}}}\par
    \endgroup
}

\begin{table*}[htbp]
\centering
\begin{minipage}{0.97\textwidth}
\raggedright
\small
\setlength{\tabcolsep}{3pt}
\renewcommand{\arraystretch}{1.16}
\skillcaseband{0.90,0.95,1.00}{\textsc{Routing Metadata and Curriculum State} ($m,\sigma$) \hfill \texttt{skill\_038}}
\vspace{0.35em}
\begin{tabular}{@{}>{\raggedright\arraybackslash\bfseries}p{0.17\linewidth} >{\raggedright\arraybackslash}p{0.79\linewidth}@{}}
    \textsc{Package} & \texttt{skill\_038} --- \texttt{book\_table\_from\_discovered\_restaurant}; added in iteration~2 with \texttt{source=skill\_induction}; \texttt{induction\_novelty=specialization}. \\
    \textsc{Description} & When a user requests a dinner reservation with a prior restaurant discovery and specifies guests, date, and time, the system should book a table at the top-rated restaurant from the discovery list. \\
    \textsc{Induction reason} & The dialogue state shows a clear pattern of task progression from restaurant discovery to table booking, grounded in user intent, prior observations, and explicit constraints. The tool selection is logically derived from the sequence of user requests and available observations, with parameters filled from prior findings and normalized into valid formats. \\
    \textsc{Statistics} & $82$ attempts; $82$ consistent; $74$ boundary and verified records; $8$ too-easy records; no too-hard or inconsistent records; mean $v_\text{solve}=0.572$; last updated in iteration~4. \\
\end{tabular}
\vspace{0.45em}
\end{minipage}
\end{table*}

\begin{table*}[htbp]
\centering
\begin{minipage}{0.97\textwidth}
\raggedright
\small
\setlength{\tabcolsep}{3pt}
\renewcommand{\arraystretch}{1.16}
\skillcaseband{1.00,0.95,0.87}{\textsc{Proposer Construction Rules} ($r$) \hfill \texttt{skill\_038}}
\vspace{0.35em}
\begin{tabular}{@{}>{\raggedright\arraybackslash\bfseries}p{0.17\linewidth} >{\raggedright\arraybackslash}p{0.79\linewidth}@{}}
    \textsc{Trigger} & The user requests a table booking after a restaurant discovery and specifies a date, time, and guest count. \\
    \textsc{Required evidence} & A prior observation contains a top-rated restaurant with an identifier, cuisine, and price constraint; the user specifies the date, time, and number of guests. \\
    \textsc{Distractors} & \texttt{find\_restaurant} and \texttt{check\_schedule} are distractors because the restaurant has already been identified. \\
    \textsc{Tool rule} & Select \texttt{book\_table} only when the user explicitly requests a table after restaurant discovery and rating. \\
    \textsc{Parameter rule} & Copy the identifier from the top-rated observation; normalize the date to \texttt{YYYY-MM-DD}, the time to \texttt{HH:MM}, and the guest count to a number. \\
    \textsc{Answer checks} & Require \texttt{restaurant\_id}, \texttt{date}, \texttt{time}, and \texttt{guests}; require valid ISO date/time formats and an identifier from the top-rated observation. \\
\end{tabular}
\vspace{0.45em}
\end{minipage}
\end{table*}

\begin{table*}[htbp]
\centering
\begin{minipage}{0.97\textwidth}
\raggedright
\small
\setlength{\tabcolsep}{3pt}
\renewcommand{\arraystretch}{1.16}
\skillcaseband{0.90,0.97,0.93}{\textsc{Proposer Hint, Validator, and Offline Oracle} ($h,\nu$) \hfill \texttt{skill\_038}}
\vspace{0.35em}
\begin{tabular}{@{}>{\raggedright\arraybackslash\bfseries}p{0.17\linewidth} >{\raggedright\arraybackslash}p{0.79\linewidth}@{}}
    \textsc{Generator hint} & Vary the city, cuisine, maximum price, date, and time while ensuring that a preceding restaurant discovery with a top rating exists. \\
    \textsc{Validator} & Tool name must be \texttt{book\_table}; \texttt{restaurant\_id} must be a string from a prior top-rated observation; \texttt{date} must use \texttt{YYYY-MM-DD}; \texttt{time} must use \texttt{HH:MM}; \texttt{guests} must be a positive integer; and all parameters must be present. \\
    \textsc{Offline recognition} & Look for a user request to book a table after a prior restaurant discovery. \\
    \textsc{Offline evidence extraction} & Extract the identifier from a top-rated observation, together with the number of guests, date, and time. \\
    \textsc{Offline decision rule} & Choose \texttt{book\_table} if a top-rated restaurant with an identifier is observed and the user requests a table booking. \\
    \textsc{Offline parameter recovery} & Fill \texttt{restaurant\_id} from the top-rated observation; normalize date to \texttt{YYYY-MM-DD}, time to \texttt{HH:MM}, and guests to an integer. \\
    \textsc{Offline final check} & Ensure that every parameter is present, valid, and derived from prior evidence. \\
\end{tabular}
\vspace{0.45em}
\end{minipage}
\end{table*}

\begin{table*}[htbp]
\centering
\begin{minipage}{0.97\textwidth}
\raggedright
\small
\setlength{\tabcolsep}{3pt}
\renewcommand{\arraystretch}{1.16}
\skillcaseband{0.95,0.95,0.95}{\textsc{Proposer Example} ($e$) \hfill \texttt{skill\_038}}
\vspace{0.35em}
\begin{tabular}{@{}>{\raggedright\arraybackslash\bfseries}p{0.17\linewidth} >{\raggedright\arraybackslash}p{0.79\linewidth}@{}}
    \textsc{Available tools} & \texttt{get\_weather(location, units)}, \texttt{find\_restaurant(city, cuisine, max\_price)}, \texttt{book\_table(restaurant\_id, date, time, guests)}, \texttt{check\_schedule(user\_email, date)}, and \texttt{send\_message(recipient, message)}. \\
    \textsc{Dialogue state} & \texttt{<user>} ``I want to plan a dinner for my friends this weekend in Portland. We're looking for a nice Italian place that doesn't cost more than \$30 per person.'' \texttt{<response>} ``I can help you find some Italian restaurants in Portland with a max price of \$30.'' \texttt{<obs>} ``Found 5 Italian restaurants in Portland with a max price of \$30.'' \texttt{<user>} ``Can you book a table for 4 of us on Saturday at 7 PM?'' \texttt{<response>} ``Which restaurant would you like me to book a table for?'' \texttt{<user>} ``Can you pick one of the top-rated ones from the list?'' \texttt{<obs>} ``Top-rated Italian restaurant in Portland with max price \$30: Trattoria al Sole, ID: \texttt{TRT-8842} (rating: $4.7$).'' \texttt{<user>} ``Based on our conversation above, please only make one tool call to solve my need.'' \\
    \textsc{Verified call} & \texttt{name=book\_table; restaurant\_id=TRT-8842; date=2023-09-16; time=19:00; guests=4}. \\
\end{tabular}
\vspace{0.45em}
\end{minipage}
\caption{\textbf{Complete induced package \texttt{skill\_038}.} The four blocks instantiate $(m,r,h,e,\nu,\sigma)$: routing metadata, construction rules, hint and validation, and a proposer example.}
\label{tab:complete_toolcall_skill}
\end{table*}

\subsubsection{Complete induced package: \texttt{skill\_042}}
\label{app:complete_toolcall_skill_timeslot}

\texttt{skill\_042} was induced in the same iteration as \texttt{skill\_038}, but captures a different dependency: a proposed tool call must follow the confirmed availability observation rather than simply mirror the user's requested time range. Its four-block listing uses the same organization as above.

\begin{table*}[htbp]
\centering
\begin{minipage}{0.97\textwidth}
\raggedright
\small
\setlength{\tabcolsep}{3pt}
\renewcommand{\arraystretch}{1.16}
\skillcaseband{0.90,0.95,1.00}{\textsc{Routing Metadata and Curriculum State} ($m,\sigma$) \hfill \texttt{skill\_042}}
\vspace{0.35em}
\begin{tabular}{@{}>{\raggedright\arraybackslash\bfseries}p{0.17\linewidth} >{\raggedright\arraybackslash}p{0.79\linewidth}@{}}
    \textsc{Package} & \texttt{skill\_042} --- \texttt{ScheduleMeetingWithTimeSlot}; added in iteration~2 with \texttt{source=skill\_induction}; \texttt{induction\_novelty=refinement}. \\
    \textsc{Description} & This pattern captures scheduling a meeting between two or more participants within a specified time window, using confirmed start and end times derived from user-provided slots and availability observations. The meeting topic is extracted directly from the user's request. \\
    \textsc{Induction reason} & The dialogue state involves a clear, structured request to schedule a meeting with specific participants, time constraints, and a topic, with the tool selection being unambiguous and grounded in explicit user input and observation. The pattern generalizes beyond this instance to any meeting scheduling request with defined participants, time window, and topic, making it reusable and nontrivial. \\
    \textsc{Statistics} & $1{,}993$ attempts; $1{,}993$ consistent; $1{,}102$ boundary and verified records; $891$ too-easy records; no too-hard or inconsistent records; mean $v_\text{solve}=0.695$; last updated in iteration~4. \\
\end{tabular}
\vspace{0.45em}
\end{minipage}
\end{table*}

\begin{table*}[htbp]
\centering
\begin{minipage}{0.97\textwidth}
\raggedright
\small
\setlength{\tabcolsep}{3pt}
\renewcommand{\arraystretch}{1.16}
\skillcaseband{1.00,0.95,0.87}{\textsc{Proposer Construction Rules} ($r$) \hfill \texttt{skill\_042}}
\vspace{0.35em}
\begin{tabular}{@{}>{\raggedright\arraybackslash\bfseries}p{0.17\linewidth} >{\raggedright\arraybackslash}p{0.79\linewidth}@{}}
    \textsc{Trigger} & The user requests a meeting between participants with a time range and topic, and an observation confirms an available time slot. \\
    \textsc{Required evidence} & The user specifies participants, a time range (for example, 10 AM to 2 PM), and a topic; an observation confirms a particular available slot within that range. \\
    \textsc{Distractors} & \texttt{SearchProduct} and \texttt{GetWeather} are irrelevant because the dialogue contains no evidence of product search, weather, or account balance. \\
    \textsc{Tool rule} & Choose \texttt{ScheduleMeeting} when the user explicitly requests a meeting with participants and a time slot, and an observation confirms availability. \\
    \textsc{Parameter rule} & Copy participants from user input; normalize \texttt{start\_time} and \texttt{end\_time} to ISO 8601 using the confirmed slot (for example, 10:30--11:30 becomes \texttt{2023-10-12T10:30:00} and \texttt{2023-10-12T11:30:00}); copy the topic from user input. \\
    \textsc{Answer checks} & Require \texttt{ScheduleMeeting}, an array of participants, ISO 8601 \texttt{start\_time} and \texttt{end\_time}, and a topic; all values must be grounded in dialogue or observation and match the schema. \\
\end{tabular}
\vspace{0.45em}
\end{minipage}
\end{table*}

\begin{table*}[htbp]
\centering
\begin{minipage}{0.97\textwidth}
\raggedright
\small
\setlength{\tabcolsep}{3pt}
\renewcommand{\arraystretch}{1.16}
\skillcaseband{0.90,0.97,0.93}{\textsc{Proposer Hint, Validator, and Offline Oracle} ($h,\nu$) \hfill \texttt{skill\_042}}
\vspace{0.35em}
\begin{tabular}{@{}>{\raggedright\arraybackslash\bfseries}p{0.17\linewidth} >{\raggedright\arraybackslash}p{0.79\linewidth}@{}}
    \textsc{Generator hint} & Vary participants (for example, John and Lisa), time ranges (for example, 9 AM to 5 PM), and topics (for example, Q3 review or budget planning), while ensuring that a confirmed time slot is provided in an observation. \\
    \textsc{Validator} & Tool name must be \texttt{ScheduleMeeting}; participants must be a non-empty string array; \texttt{start\_time} and \texttt{end\_time} must be ISO 8601 timestamps; start must precede end; both timestamps must lie within the requested window; the topic must be non-empty; and no parameter may be invented. \\
    \textsc{Offline recognition} & Look for requests involving ``schedule a meeting'' or ``plan a meeting'' with participants and time constraints. \\
    \textsc{Offline evidence extraction} & Extract the participants, the requested start and end range, and the topic. \\
    \textsc{Offline decision rule} & If an observation confirms a specific available slot within the requested window, select \texttt{ScheduleMeeting}. \\
    \textsc{Offline parameter recovery} & Normalize times to ISO 8601; copy participants and topic directly; use the confirmed slot as the start and end times. \\
    \textsc{Offline final check} & Ensure that all parameters are present, valid, derived from dialogue or observation, and free of schema violations. \\
\end{tabular}
\vspace{0.45em}
\end{minipage}
\end{table*}

\begin{table*}[htbp]
\centering
\begin{minipage}{0.97\textwidth}
\raggedright
\small
\setlength{\tabcolsep}{3pt}
\renewcommand{\arraystretch}{1.16}
\skillcaseband{0.95,0.95,0.95}{\textsc{Proposer Example} ($e$) \hfill \texttt{skill\_042}}
\vspace{0.35em}
\begin{tabular}{@{}>{\raggedright\arraybackslash\bfseries}p{0.17\linewidth} >{\raggedright\arraybackslash}p{0.79\linewidth}@{}}
    \textsc{Available tools} & \texttt{GetWeather(location, units)} with \texttt{units} in \{\texttt{celsius}, \texttt{fahrenheit}\}; \texttt{SearchProduct(query, category)} with \texttt{category} in \{\texttt{electronics}, \texttt{clothing}, \texttt{books}, \texttt{home}\}; \texttt{ScheduleMeeting(participants, start\_time, end\_time, topic)}; \texttt{GetAccountBalance(account\_id)}; \texttt{SendEmail(to, subject, body)}; and \texttt{FindRoute(origin, destination, mode)} with \texttt{mode} in \{\texttt{driving}, \texttt{walking}, \texttt{cycling}\}. \\
    \textsc{Dialogue state} & \texttt{<user>} ``I need to plan a meeting between Sarah and Alex next week. The meeting should be between 10 AM and 2 PM on Thursday, and the topic is project updates.'' \texttt{<response>} ``I can help you schedule that meeting. Could you confirm the participants and preferred time slot?'' \texttt{<user>} ``Yes, participants are Sarah and Alex. Confirm the meeting on Thursday, 10 AM to 2 PM.'' \texttt{<response>} ``I'll check if that time works. I need to schedule the meeting now.'' \texttt{<obs>} ``Available time slots for Sarah and Alex on Thursday between 10:00 and 14:00 are fully booked except for 10:30 to 11:30.'' \texttt{<user>} ``That works. Let's go with 10:30 to 11:30.'' \texttt{<response>} ``Great! I'll proceed with scheduling the meeting at 10:30 AM to 11:30 AM on Thursday for Sarah and Alex on the topic of project updates.'' Final request: ``Based on our conversation above, please only make one tool call to solve my need.'' \\
    \textsc{Verified call} & \texttt{name=ScheduleMeeting; participants=[Sarah, Alex]; start\_time=2023-10-12T10:30:00; end\_time=2023-10-12T11:30:00; topic=project updates}. \\
\end{tabular}
\vspace{0.45em}
\end{minipage}
\caption{\textbf{Complete induced package \texttt{skill\_042}.} The four blocks instantiate $(m,r,h,e,\nu,\sigma)$: routing metadata, construction rules, hint and validation, and a proposer example.}
\label{tab:complete_toolcall_skill_timeslot}
\end{table*}

These examples are more structured than a direct keyword-to-tool association. They bind the current action to evidence accumulated in earlier turns: a selected candidate becomes an identifier, and an observed slot becomes the executable interval. The corresponding package statistics show that the patterns are not merely retained as descriptions: both contributed filtered frontier records during routing. More broadly, the induced library contains analogous search-to-action and context-to-parameter patterns across restaurant, calendar, flight, and reminder workflows. This qualitative trace supports the mechanism underlying Skill-SP's evolving curriculum coverage: exploration can be consolidated into reusable, verifiable task patterns.

\subsection{Logical Reasoning}
\label{app:logical_skill_cases}

Logical-reasoning packages encode reusable constraint topologies rather than tool-selection rules. Each package combines a procedural construction rule with a declarative compiler specification that guides the fixed logical runtime in assembling constraint blueprints. The runtime then admits an instance only when its schema, constraints, and unique solution pass deterministic checks. The exported logical library does not populate per-package routing statistics; accordingly, we report saved accepted structural-evidence records below rather than treating these counts as solver performance.

\begin{table*}[htbp]
    \centering
    \small
    \renewcommand{\arraystretch}{1.22}
    \setlength{\tabcolsep}{3.5pt}
    \begin{tabular*}{\textwidth}{@{\extracolsep{\fill}} >{\raggedright\arraybackslash}p{0.17\textwidth} >{\raggedright\arraybackslash}p{0.28\textwidth} >{\raggedright\arraybackslash}p{0.28\textwidth} >{\raggedright\arraybackslash}p{0.19\textwidth}}
        \toprule
        \textbf{Induced package} & \textbf{Constraint topology} & \textbf{Compiler-enforced structure} & \textbf{Saved structural evidence} \\
        \midrule
        \texttt{zebra\_skill\_022}\newline
        \emph{Fixed-anchor adjacency bridge}\newline
        Iteration 1
        & A fixed entity, a cross-attribute \texttt{next\_to} relation, and a \texttt{same\_house} bridge resolve a positional ambiguity.
        & \texttt{anchored\_component} $\rightarrow$ \texttt{component\_bridge} $\rightarrow$ \texttt{same\_house\_chain}; requires at least one \texttt{fixed}, \texttt{same\_house}, and \texttt{next\_to} clue.
        & $333$ accepted and $4$ rejected evidence records. \\
        \addlinespace[0.45em]
        \texttt{zebra\_skill\_031}\newline
        \emph{Chain-anchor-bridge}\newline
        Iteration 2
        & A directed three-value chain is anchored at one endpoint, then connected across attributes through \texttt{same\_house} and \texttt{next\_to} bridges.
        & \texttt{directed\_chain} $\rightarrow$ \texttt{same\_house\_chain} $\rightarrow$ \texttt{component\_bridge}; requires at least two \texttt{directly\_left\_of}, one \texttt{same\_house}, and one \texttt{next\_to} clue.
        & $178$ accepted and $2$ rejected evidence records. \\
        \bottomrule
    \end{tabular*}
    \caption{\textbf{Induced logical-reasoning skill patterns.} Both packages are saved with \texttt{source=zebra\_skill\_induction}. Counts are accepted or rejected structural-evidence records, not solver scores. Every accepted record reported here has valid schema and constraints, a unique puzzle, and a capped solution count of one.}
    \label{tab:logical_skill_cases}
\end{table*}

\subsubsection{Complete induced package: \texttt{zebra\_skill\_022}}
\label{app:complete_logical_skill_anchor}

We first expand \texttt{zebra\_skill\_022}, an early-stage fixed-anchor adjacency bridge induced in iteration~1. A fixed entity, a cross-attribute \texttt{next\_to} relation, and a \texttt{same\_house} component jointly resolve a non-local positional dependency. The following five blocks give its complete package-specific content using the $(m,r,h,e,\nu,\sigma)$ organization.

\begin{table*}[htbp]
\centering
\begin{minipage}{0.97\textwidth}
\raggedright
\small
\setlength{\tabcolsep}{3pt}
\renewcommand{\arraystretch}{1.16}
\skillcaseband{0.90,0.95,1.00}{\textsc{Routing Metadata and Induction Record} ($m,\sigma$) \hfill \texttt{zebra\_skill\_022}}
\vspace{0.35em}
\begin{tabular}{@{}>{\raggedright\arraybackslash\bfseries}p{0.17\linewidth} >{\raggedright\arraybackslash}p{0.79\linewidth}@{}}
    \textsc{Package} & \texttt{zebra\_skill\_022} --- \texttt{Fixed-Anchor Adjacency Bridge}; added in iteration~1 with \texttt{source=zebra\_skill\_induction}. \\
    \textsc{Description} & A \texttt{next\_to} relation links a fixed-position entity to a value in another attribute, while a \texttt{same\_house} component anchors a value pair. Together, the fixed anchor and adjacency bridge resolve a non-local positional dependency. \\
    \textsc{Induction reason} & The inducing trace contained directed chains and cross-attribute order relations but no reusable pattern combining a fixed entity, cross-attribute adjacency, and a \texttt{same\_house} anchor. This package adds that missing adjacency-based bridge. \\
    \textsc{Induction novelty} & The bridge resolves relative positions through both adjacency and shared-house structure, rather than isolated \texttt{left\_of} or \texttt{directly\_left\_of} constraints. \\
    \textsc{Evidence state} & The exported \texttt{stats.json} has no populated routing counters (the $\sigma$ fields are zero). Independently saved refinement evidence contains $333$ accepted and $4$ rejected records; each accepted record satisfies schema validity, constraint validity, uniqueness, and capped solution count. \\
\end{tabular}
\vspace{0.45em}
\end{minipage}
\end{table*}

\begin{table*}[htbp]
\centering
\begin{minipage}{0.97\textwidth}
\raggedright
\small
\setlength{\tabcolsep}{3pt}
\renewcommand{\arraystretch}{1.16}
\skillcaseband{1.00,0.95,0.87}{\textsc{Proposer Construction Resource} ($r$): Declarative Compiler \hfill \texttt{zebra\_skill\_022}}
\vspace{0.35em}
\begin{tabular}{@{}>{\raggedright\arraybackslash\bfseries}p{0.17\linewidth} >{\raggedright\arraybackslash}p{0.79\linewidth}@{}}
    \textsc{Compiler spec} & \texttt{version=1}; \texttt{fixed\_limit=1}; preferred relation types are \texttt{fixed}, \texttt{same\_house}, and \texttt{next\_to}. \\
    \textsc{Seed program} & (1) \texttt{anchored\_component(anchors=1, component\_width=2)}; (2) \texttt{component\_bridge(component\_width=2, relation=next\_to)}; (3) \texttt{same\_house\_chain(width=2)}. \\
    \textsc{Compiler guards} & \texttt{prune=true} and \texttt{ensure\_relation\_diversity=true}; \texttt{derived\_from=zebra\_skill\_010}. \\
\end{tabular}
\vspace{0.45em}
\end{minipage}
\end{table*}

\begin{table*}[htbp]
\centering
\begin{minipage}{0.97\textwidth}
\raggedright
\small
\setlength{\tabcolsep}{3pt}
\renewcommand{\arraystretch}{1.16}
\skillcaseband{1.00,0.95,0.87}{\textsc{Proposer Construction Rules} ($r$) \hfill \texttt{zebra\_skill\_022}}
\vspace{0.35em}
\begin{tabular}{@{}>{\raggedright\arraybackslash\bfseries}p{0.17\linewidth} >{\raggedright\arraybackslash}p{0.79\linewidth}@{}}
    \textsc{Generation trigger} & Use this pattern when a fixed-position entity is linked by \texttt{next\_to} to a value in another attribute and a \texttt{same\_house} constraint anchors a value pair. \\
    \textsc{Construction rule} & Start from a fixed value in a specific house. Add a cross-attribute \texttt{next\_to} relation, then a \texttt{same\_house} pair in a different house so that the adjacency and shared-house components jointly resolve positions. \\
    \textsc{Difficulty controls} & Increase difficulty by using a non-extreme fixed house, adding a \texttt{not\_house} exclusion that blocks one adjacent position, or choosing a non-adjacent \texttt{same\_house} component. \\
\end{tabular}
\vspace{0.45em}
\end{minipage}
\end{table*}

\begin{table*}[htbp]
\centering
\begin{minipage}{0.97\textwidth}
\raggedright
\small
\setlength{\tabcolsep}{3pt}
\renewcommand{\arraystretch}{1.16}
\skillcaseband{0.90,0.97,0.93}{\textsc{Proposer Hint and Package Validator} ($h,\nu$) \hfill \texttt{zebra\_skill\_022}}
\vspace{0.35em}
\begin{tabular}{@{}>{\raggedright\arraybackslash\bfseries}p{0.17\linewidth} >{\raggedright\arraybackslash}p{0.79\linewidth}@{}}
    \textsc{Generator hint} & Anchor a fixed value in a specific house; connect a name to a hobby or sport through \texttt{next\_to}; and add a \texttt{same\_house} pair for two attributes. Keep the \texttt{next\_to} relation adjacent to the fixed entity, and avoid placing the \texttt{same\_house} pair in the fixed entity's house unless required. \\
    \textsc{Type guard} & Require at least one \texttt{fixed}, one \texttt{same\_house}, and one \texttt{next\_to} relation. \\
    \textsc{Local validator} & Verify the fixed house, adjacency, and shared-house relation; require the resulting constraint system to have a unique solution. \\
    \textsc{Offline oracle} & Assign the fixed value to its house; enumerate adjacent houses for the \texttt{next\_to} value; apply the \texttt{same\_house} pair; eliminate fixed-anchor and adjacency conflicts; then solve the remaining CSP under all-different and adjacency constraints. \\
\end{tabular}
\vspace{0.45em}
\end{minipage}
\end{table*}

\begin{table*}[htbp]
\centering
\begin{minipage}{0.97\textwidth}
\raggedright
\small
\setlength{\tabcolsep}{3pt}
\renewcommand{\arraystretch}{1.16}
\skillcaseband{0.95,0.95,0.95}{\textsc{Proposer Example} ($e$) and Accepted Offline Instance \hfill \texttt{zebra\_skill\_022}}
\vspace{0.35em}
\begin{tabular}{@{}>{\raggedright\arraybackslash\bfseries}p{0.17\linewidth} >{\raggedright\arraybackslash}p{0.79\linewidth}@{}}
    \textsc{Stored package example} & Two houses; \texttt{Name}=\{Elena, Felix\}, \texttt{Nationality}=\{German, Japanese\}, \texttt{Drink}=\{Coffee, Tea\}, \texttt{Hobby}=\{Gardening, Reading\}, and \texttt{FavoriteSport}=\{Badminton, Soccer\}. Clues: (1) Elena directly left of Felix; (2) Felix in House 2; (3) Elena left of Felix; (4) German same house as Soccer; (5) Tea left of Coffee; (6) Elena next to Badminton; (7) Gardening not in House 1. \\
    \textsc{Example rationale} & Felix fixed in House 2 and Elena next to Badminton force Elena to House 1 and Badminton to House 2; the \texttt{same\_house} component then resolves the remaining positional bridge. \\
    \textsc{Accepted blueprint} & Three houses; \texttt{Name}=\{Leo, Mia, Ryan\}, \texttt{Nationality}=\{French, Indian, Swedish\}, \texttt{Drink}=\{Latte, GreenTea, Cappuccino\}, \texttt{Hobby}=\{Photography, Cooking, Reading\}, and \texttt{FavoriteSport}=\{Tennis, Badminton, Volleyball\}; difficulty \texttt{medium}; theme \emph{City Apartment Residents}. \\
    \textsc{Compiled clues} & (1) Leo in House 2; (2) Leo same house as Indian; (3) Mia same house as Swedish; (4) Indian same house as Cappuccino; (5) Swedish next to Cappuccino; (6) Ryan same house as French; (7) GreenTea same house as Volleyball; (8) Swedish next to Tennis; (9) Photography left of Reading; (10) Volleyball directly left of Indian; (11) French same house as Cooking; (12) Photography left of Cooking. \\
    \textsc{Unique solution} & House 1: GreenTea, Volleyball, Photography, Mia, Swedish; House 2: Cappuccino, Tennis, Reading, Leo, Indian; House 3: Latte, Badminton, Cooking, Ryan, French. \\
    \textsc{Structural checks} & \texttt{logical\_generation\_mode=blueprint}; \texttt{prompt\_valid=1}; \texttt{schema\_valid=1}; \texttt{constraints\_valid=1}; \texttt{puzzle\_unique=1}; \texttt{solution\_count\_capped=1}. \\
\end{tabular}
\vspace{0.45em}
\end{minipage}
\caption{\textbf{Complete induced package \texttt{zebra\_skill\_022}.} The five blocks instantiate $(m,r,h,e,\nu,\sigma)$: metadata, compiler resource, construction rules, hint and validation, and a stored example with an accepted offline instance.}
\label{tab:complete_logical_skill_anchor}
\end{table*}

\subsubsection{Complete induced package: \texttt{zebra\_skill\_031}}
\label{app:complete_logical_skill}

We next expand \texttt{zebra\_skill\_031}, a later chain--anchor--bridge pattern induced in iteration~2. It extends the fixed-anchor construction above with a directed chain and two cross-attribute bridge types. The following five blocks expose its package fields using the structure defined in Section~\ref{sec:proposer_optimization}, together with one accepted unique puzzle instance.

\begin{table*}[htbp]
\centering
\begin{minipage}{0.97\textwidth}
\raggedright
\small
\setlength{\tabcolsep}{3pt}
\renewcommand{\arraystretch}{1.16}
\skillcaseband{0.90,0.95,1.00}{\textsc{Routing Metadata and Induction Record} ($m,\sigma$) \hfill \texttt{zebra\_skill\_031}}
\vspace{0.35em}
\begin{tabular}{@{}>{\raggedright\arraybackslash\bfseries}p{0.17\linewidth} >{\raggedright\arraybackslash}p{0.79\linewidth}@{}}
    \textsc{Package} & \texttt{zebra\_skill\_031} --- \texttt{Chain-Anchor-Bridge Pattern}; added in iteration~2 with \texttt{source=zebra\_skill\_induction}; \texttt{induction\_novelty} is a hybrid chain-solving structure. \\
    \textsc{Description} & A directed adjacency chain is anchored by a fixed value and reinforced with cross-attribute \texttt{same\_house} and \texttt{next\_to} bridges, so deduction propagates through value-sharing and adjacency rather than position alone. \\
    \textsc{Induction reason} & The inducing trace contains a linear \texttt{directly\_left\_of} chain (Elara--Gita--Leo), anchored by Leo in House 3 and extended through cross-attribute links. The pattern adds \texttt{same\_house} and \texttt{next\_to} bridges that the source directed-chain skill does not model. \\
    \textsc{Evidence summary} & The exported \texttt{stats.json} has no populated routing counters (the $\sigma$ fields are zero). The saved refinement evidence contains $178$ accepted and $2$ rejected records. Every accepted record has \texttt{constraints\_valid=1}, \texttt{schema\_valid=1}, \texttt{puzzle\_unique=1}, and \texttt{solution\_count\_capped=1}. \\
\end{tabular}
\vspace{0.45em}
\end{minipage}
\end{table*}

\begin{table*}[htbp]
\centering
\begin{minipage}{0.97\textwidth}
\raggedright
\small
\setlength{\tabcolsep}{3pt}
\renewcommand{\arraystretch}{1.16}
\skillcaseband{1.00,0.95,0.87}{\textsc{Proposer Construction Resource} ($r$): Declarative Compiler \hfill \texttt{zebra\_skill\_031}}
\vspace{0.35em}
\begin{tabular}{@{}>{\raggedright\arraybackslash\bfseries}p{0.17\linewidth} >{\raggedright\arraybackslash}p{0.79\linewidth}@{}}
    \textsc{Runtime spec} & \texttt{version=1}; \texttt{fixed\_limit=1}; preferred relation types are \texttt{directly\_left\_of}, \texttt{same\_house}, and \texttt{next\_to}. \\
    \textsc{Seed program} & (1) \texttt{directed\_chain(length=3, anchor\_end=true)}; (2) \texttt{same\_house\_chain(width=2)}; (3) \texttt{component\_bridge(component\_width=2, relation=next\_to)}. \\
    \textsc{Compiler guards} & \texttt{prune=true} and \texttt{ensure\_relation\_diversity=true}. \\
    \textsc{Lineage} & \texttt{derived\_from=zebra\_skill\_002}. \\
\end{tabular}
\vspace{0.45em}
\end{minipage}
\end{table*}

\begin{table*}[htbp]
\centering
\begin{minipage}{0.97\textwidth}
\raggedright
\small
\setlength{\tabcolsep}{3pt}
\renewcommand{\arraystretch}{1.16}
\skillcaseband{1.00,0.95,0.87}{\textsc{Proposer Construction Rules} ($r$) \hfill \texttt{zebra\_skill\_031}}
\vspace{0.35em}
\begin{tabular}{@{}>{\raggedright\arraybackslash\bfseries}p{0.17\linewidth} >{\raggedright\arraybackslash}p{0.79\linewidth}@{}}
    \textsc{Generation trigger} & Use this pattern when a directed adjacency chain is anchored by a fixed entity and must be resolved through cross-attribute \texttt{same\_house} or \texttt{next\_to} constraints that link values across chain members. \\
    \textsc{Construction rule} & Begin with a directed chain of at least three \texttt{directly\_left\_of} relations anchored at one end by \texttt{fixed}. Add non-redundant, non-symmetric \texttt{same\_house} or \texttt{next\_to} bridges that link a chain member to another value or a fixed house. \\
    \textsc{Offline oracle} & Resolve positional offsets from the anchor; then apply \texttt{same\_house} or \texttt{next\_to} to determine value assignments across the chain. Propagate shared or adjacent values, and resolve conflicts using all-different and cross-attribute consistency. \\
    \textsc{Difficulty controls} & Extend the chain to four values, add multiple bridges, or place bridges between non-consecutive members. Avoid symmetric or isolated constraints. \\
\end{tabular}
\vspace{0.45em}
\end{minipage}
\end{table*}

\begin{table*}[htbp]
\centering
\begin{minipage}{0.97\textwidth}
\raggedright
\small
\setlength{\tabcolsep}{3pt}
\renewcommand{\arraystretch}{1.16}
\skillcaseband{0.90,0.97,0.93}{\textsc{Proposer Hint and Package Validator} ($h,\nu$) \hfill \texttt{zebra\_skill\_031}}
\vspace{0.35em}
\begin{tabular}{@{}>{\raggedright\arraybackslash\bfseries}p{0.17\linewidth} >{\raggedright\arraybackslash}p{0.79\linewidth}@{}}
    \textsc{Generator hint} & Anchor a directed chain of length three or four with a fixed entity. Add a \texttt{same\_house} or \texttt{next\_to} bridge to a non-trivial attribute such as nationality or hobby, rather than only a name or drink. Avoid symmetric or isolated bridges. \\
    \textsc{Local validator} & Require \texttt{directly\_left\_of}, \texttt{same\_house}, and \texttt{next\_to}, with minimum counts $2$, $1$, and $1$, respectively. Every constraint must hold, the bridge logic must yield a unique solution, and at least one non-redundant \texttt{same\_house} or \texttt{next\_to} connection must link to a chain member. \\
    \textsc{Offline filtering oracle} & (1) Resolve the directed chain from the anchor. (2) Apply \texttt{same\_house} or \texttt{next\_to} to determine value positions. (3) Propagate values through shared attributes. (4) Use adjacency around known values. (5) Apply all-different elimination. (6) Validate uniqueness by full constraint satisfaction. \\
\end{tabular}
\vspace{0.45em}
\end{minipage}
\end{table*}

\begin{table*}[htbp]
\centering
\begin{minipage}{0.97\textwidth}
\raggedright
\small
\setlength{\tabcolsep}{3pt}
\renewcommand{\arraystretch}{1.16}
\skillcaseband{0.95,0.95,0.95}{\textsc{Proposer Example} ($e$) and Accepted Offline Instance \hfill \texttt{zebra\_skill\_031}}
\vspace{0.35em}
\begin{tabular}{@{}>{\raggedright\arraybackslash\bfseries}p{0.17\linewidth} >{\raggedright\arraybackslash}p{0.79\linewidth}@{}}
    \textsc{Package example} & Chain: Elara $\rightarrow$ Gita $\rightarrow$ Leo, with Leo fixed in House 3. Cross-attribute bridges connect Cooking to Mia, Mia to Reading through \texttt{same\_house}, and Brazilian to Coffee through \texttt{next\_to}. The resulting propagation resolves Mia, Reading, Brazilian, Coffee, and Tea without relying only on positional logic. \\
    \textsc{Accepted blueprint} & Six houses; attributes \texttt{Name} = \{Clara, Dale, Eva, Finn, Gia, Leo\} and \texttt{Hobby} = \{Reading, Cooking, Gardening, Photography, Painting, Writing\}; difficulty \texttt{medium}; theme \emph{Summer Reading Club}. \\
    \textsc{Compiled clues} & (1) Dale directly left of Gia; (2) Gia directly left of Leo; (3) Leo in House 6; (4) Eva same house as Cooking; (5) Eva next to Gardening; (6) Finn directly left of Dale; (7) Reading next to Gia; (8) Gardening left of Painting; (9) Gardening directly left of Dale; (10) Leo same house as Photography. \\
    \textsc{Unique solution} & House 1: Clara, Writing; House 2: Eva, Cooking; House 3: Finn, Gardening; House 4: Dale, Reading; House 5: Gia, Painting; House 6: Leo, Photography. \\
    \textsc{Structural checks} & \texttt{logical\_generation\_mode=blueprint}; \texttt{prompt\_valid=1}; \texttt{schema\_valid=1}; \texttt{constraints\_valid=1}; \texttt{puzzle\_unique=1}; \texttt{solution\_count\_capped=1}. \\
\end{tabular}
\vspace{0.45em}
\end{minipage}
\caption{\textbf{Complete induced package \texttt{zebra\_skill\_031}.} The five blocks instantiate $(m,r,h,e,\nu,\sigma)$: metadata, compiler resource, construction rules, hint and validation, and a package example with an accepted offline instance.}
\label{tab:complete_logical_skill}
\end{table*}

The logical examples illustrate a different form of library evolution from the tool-call setting. Induction adds reusable constraint topologies and their compiler constraints, enabling subsequent generations to jointly structure directed chains, cross-attribute bridges, and uniqueness conditions.